%% file: main.tex
\PassOptionsToPackage{table}{xcolor}
\documentclass{style}
\usepackage[T1]{fontenc}
\usepackage{amsmath,amssymb}
\usepackage{graphicx}
\usepackage{booktabs}
\usepackage{multirow}
\usepackage{array}
\usepackage{enumitem}
\usepackage{fontawesome5}
\usepackage{wrapfig}
\usepackage{needspace}
\usepackage{microtype}
\usepackage[scaled=0.92]{zi4}
\usepackage{xspace}
\makeatletter
\renewcommand{\BreakableUnderscore}{\leavevmode\prw@zbreak\textunderscore\discretionary{}{}{}\prw@zbreak}
\makeatother
\setlist{leftmargin=*,itemsep=1pt,topsep=2pt,parsep=0pt}
\input{math_commands.tex}

\input{macro.tex}
\hypersetup{
  pdftitle={See What I Need, Dream What I Want: A Navigation Foundation Model on a Self-Constructed Representation Space},
  pdfauthor={Jiawei Chen, Yinxi Wang, Zihan Yang, Meijun Gao, Yaowei Liu, Shuming Zhang, Zecheng Li, Xiangju Mi, Yuxuan Liu, Runzhong Huang, Tianlin Li, Xun Zhang, Zhengyang Wang, Xu Miao, Tong Yang}}
\title{See What I Need, Dream What I Want:\\
A Navigation Foundation Model on a Self-Constructed Representation Space}
\renewcommand{\authorfont}{\fontsize{11.5}{14}\selectfont}
\author{
  Jiawei~Chen\texorpdfstring{$^{*1}$}{} \hspace{0.1cm}
  Yinxi~Wang\texorpdfstring{$^{*\dagger1}$}{} \hspace{0.1cm}
  Zihan~Yang\texorpdfstring{$^{*1}$}{} \hspace{0.1cm}
  Meijun~Gao\texorpdfstring{$^{*1}$}{} \hspace{0.1cm}
  Yaowei~Liu\texorpdfstring{$^{*1}$}{}\\
  Shuming~Zhang\texorpdfstring{$^{*1}$}{} \hspace{0.1cm}
  Zecheng~Li\texorpdfstring{$^{2}$}{} \hspace{0.1cm}
  Xiangju~Mi\texorpdfstring{$^{1}$}{} \hspace{0.1cm}
  Yuxuan~Liu\texorpdfstring{$^{1}$}{} \hspace{0.1cm}
  Runzhong~Huang\texorpdfstring{$^{1}$}{}\\
  Tianlin~Li\texorpdfstring{$^{1}$}{} \hspace{0.1cm}
  Xun~Zhang\texorpdfstring{$^{1}$}{} \hspace{0.1cm}
  Zhengyang~Wang\texorpdfstring{$^{1}$}{} \hspace{0.1cm}
  Xu~Miao\texorpdfstring{$^{1,3\,\text{\faIcon[regular]{envelope}}}$}{} \hspace{0.1cm}
  Tong~Yang\texorpdfstring{$^{1\,\text{\faIcon[regular]{envelope}}}$}{}
}
\affiliation[1]{Peking University}
\affiliation[2]{University College London, University of London}
\affiliation[3]{DataCanvas}
\contribution[*]{Equal contribution}
\contribution[\dagger]{Project lead}
\contribution[\text{\faIcon[regular]{envelope}}]{Corresponding authors}
\checkdata[
\raisebox{-0.2em}{\includegraphics[width=0.025\linewidth]{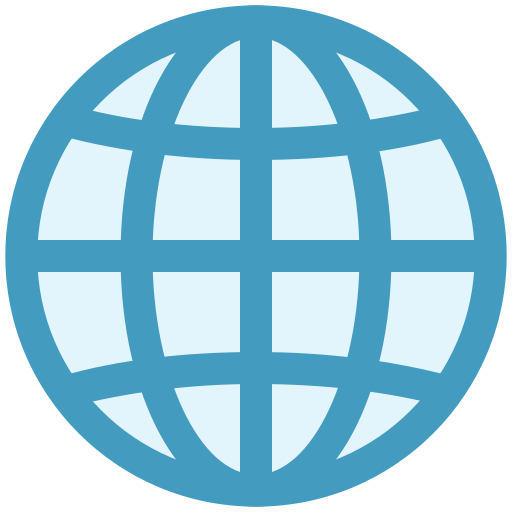}}~~Project Page]{\href{https://wang-yinxi.github.io/representnav/}{\texttt{https://wang-yinxi.github.io/representnav/}}
\\[-1.5ex]}
\checkdata[
\raisebox{-0.2em}{\includegraphics[width=0.025\linewidth]{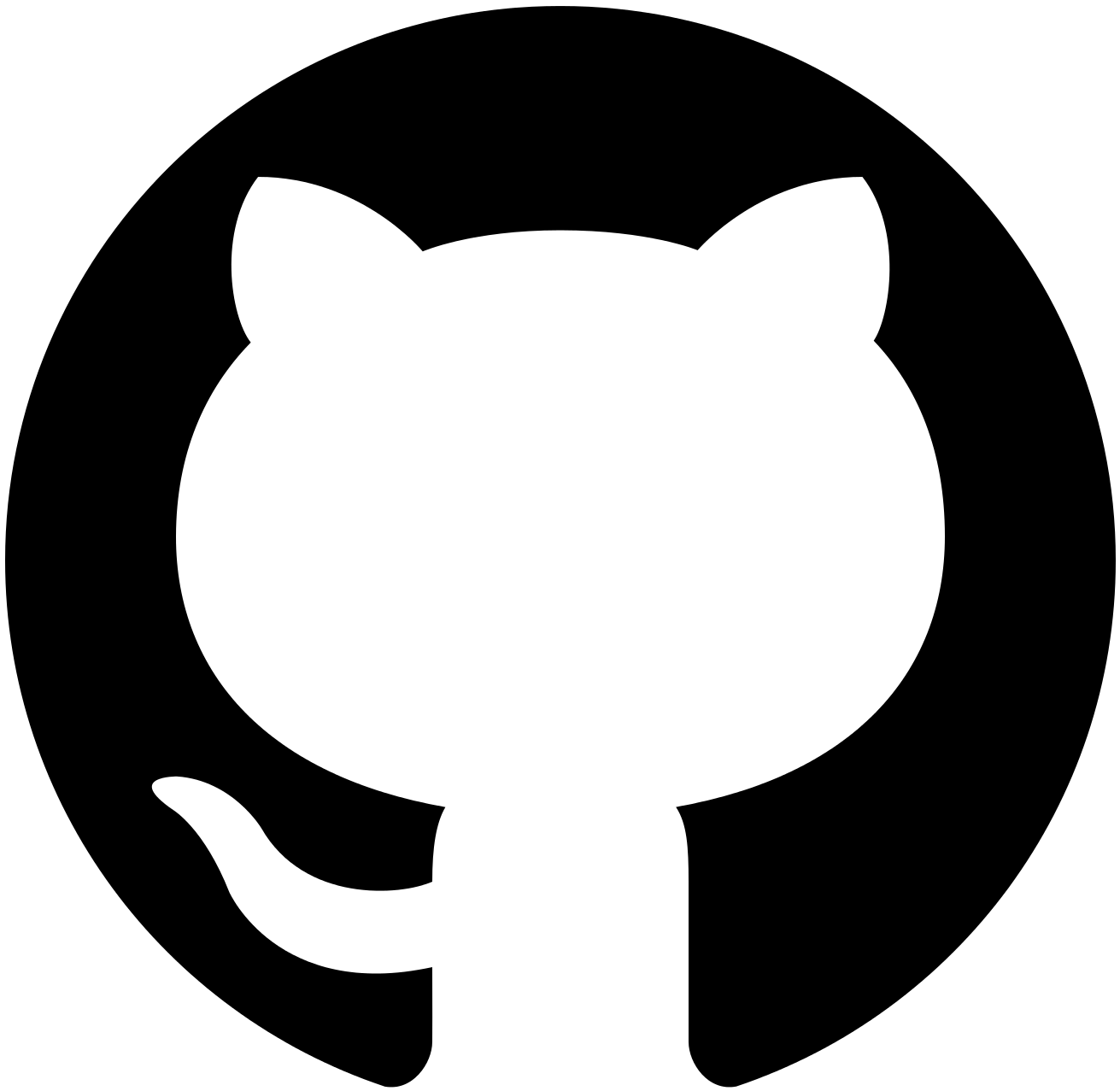}}~~GitHub Repo]{\href{https://github.com/wang-yinxi/RepresentNav}{\texttt{https://github.com/wang-yinxi/RepresentNav}}
\\[-1.7ex]}
\checkdata[
\raisebox{-0.2em}{\includegraphics[width=0.025\linewidth]{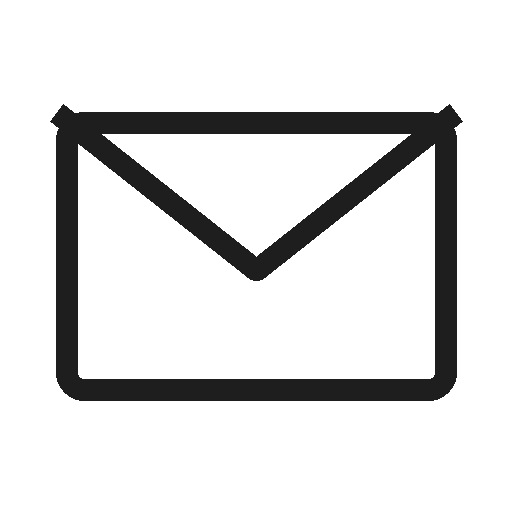}}~~Corresponding Authors]{\textcolor{pkured}{\texttt{Xu Miao, Tong Yang}}
\\[-1.7ex]}
\abstract{\input{section/abstract}}
\begin{document}
\maketitle
\suppressfloats[t]
\begin{figure}[t!]
\centering
\includegraphics[width=\linewidth,trim=0 32pt 0 34pt,clip]{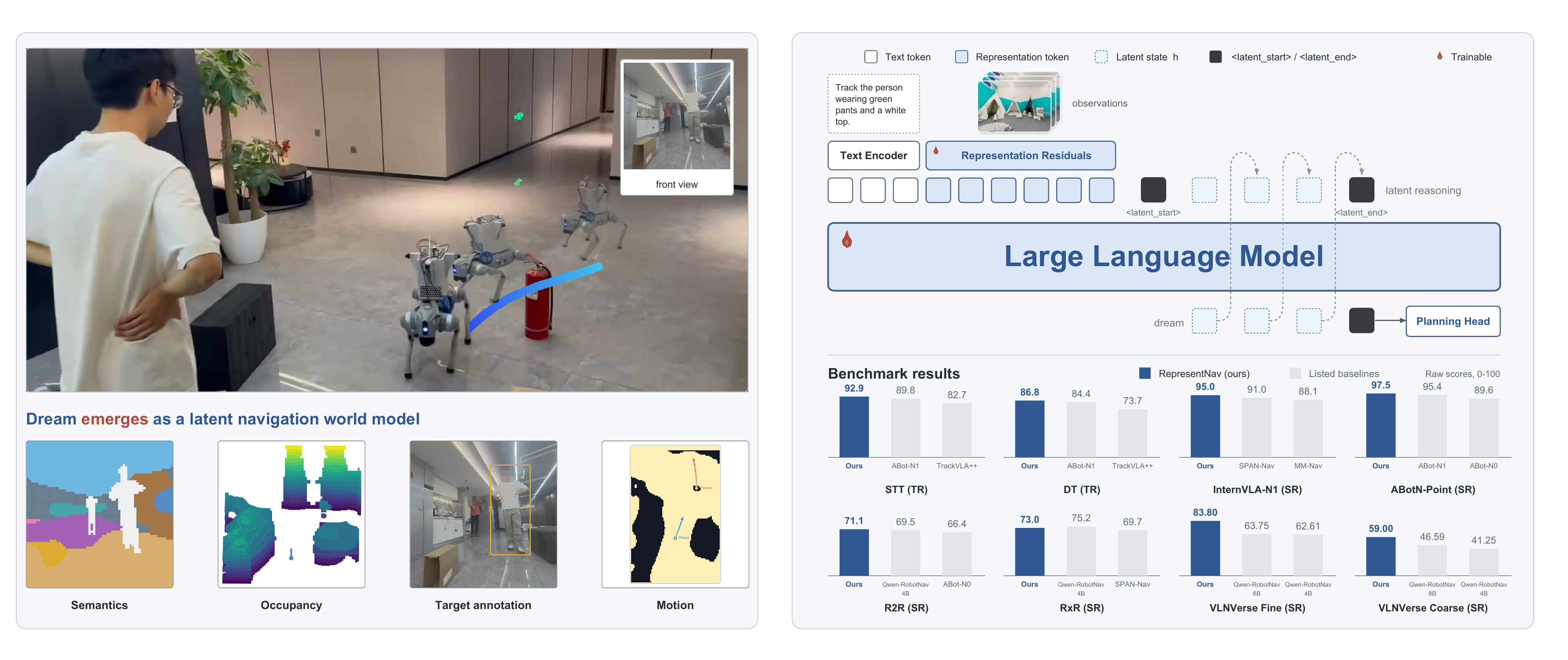}
\caption{\textbf{See what I need, dream what I want.} \ours learns a shared visual interface over frozen visual priors and rolls out latent predictions before generating waypoints. Left: real-world person tracking and post-hoc readouts of semantics, occupancy, and motion. Right: model overview and benchmark results.}
\label{fig:representnav_teaser}
\end{figure}
\input{section/introduction}
\input{section/related_works}
\input{section/method}

\input{section/experiments}
\input{section/analysis}

\input{section/conclusion}
\clearpage
\bibliographystyle{plainnat}
\setlength{\bibhang}{0pt}
\setlength\bibindent{0pt}
\bibliography{iclr2027_conference,representnav}
\clearpage
\appendix
{\centering
\textbf{\Large See What I Need, Dream What I Want: A Navigation Foundation Model on a Self-Constructed Representation Space}\par
\vspace{5pt}
{\Large Appendix}
\par}
\input{section/appendix}
\end{document}

%% file: math_commands.tex
\usepackage{amsmath,amsfonts,bm}

\def\eqref#1{equation~\ref{#1}}

\def\1{\bm{1}}

\DeclareMathAlphabet{\mathsfit}{\encodingdefault}{\sfdefault}{m}{sl}
\SetMathAlphabet{\mathsfit}{bold}{\encodingdefault}{\sfdefault}{bx}{n}



%% file: macro.tex
\usepackage{amssymb}
\def\ours{RepresentNav\xspace}
\providecommand{\mixres}{Representation Residuals\xspace}
\newcommand{\todo}[1]{}
\newcommand{\xiangju}[1]{}
\newcommand{\jiawei}[1]{}
\newlength{\capline}

\newcommand{\subtab}[1]{\par\noindent{\sffamily\bfseries\fontsize{8.5}{10.5}\selectfont\color{pkured}#1}\par\vspace{3pt}}

%% file: section/abstract.tex
What visual representation should a navigation policy act on? A central idea of representation learning is that the task's own error should shape the representation. Navigation policies built on vision foundation models largely set this idea aside: the encoders' pretraining objectives decide much of what the vision-language backbone sees. We present \ours, a navigation foundation model built on a \emph{self-constructed representation space}: a visual representation constructed by the policy's own objectives out of the internal structure of vision foundation models. \ours builds this space with \mixres, a compact three-layer transformer that attends over features spanning the full depth of multiple complementary vision foundation models, so that the policy learns to see what it needs for navigation. In the same space it dreams what it wants: before every action a latent chain-of-thought predicts the representation of the scene a moment ahead. The dream is trained end-to-end with the policy as a joint-embedding predictive architecture (JEPA), with a next-representation prediction loss and a regularizer on its targets. \ours sets a new state of the art on VLNVerse, VLN-CE R2R, EVT-Bench, and two point-goal benchmarks, and its gains carry over to a physical robot amid clutter and moving people. On the VLNVerse fine-grained benchmark, \ours achieves a success rate of 83.80\%, outperforming the previous state of the art by 20.05 percentage points, and on EVT-Bench distracted tracking it reaches 85.5\% with a single camera (+10.1). Analysis shows where they come from. With the same encoders, the self-constructed space clearly outperforms concatenating their outputs and learns to weight their layers unevenly, and a dream in it cuts collisions by a quarter. A single dreamed latent decodes into the depth, semantics, and 3D occupancy of the scene a moment ahead, more accurately than pretrained encoders applied to the real future frame, although occupancy is never supervised, and the rollout tracks the robot's own motion over the next second: the dream emerges as a compact latent world model.

%% file: section/introduction.tex
\section{Introduction}\label{sec:intro}

Consider a robot asked to cross a cluttered hall to a point on its far side. What it needs to see is not every pixel but a representation: a description of the world that makes explicit what the next decision needs and leaves out the rest \citep{bengio2013representation}. A person may name three things, how far the nearest surfaces are, what the objects around it are, and how those objects and the robot itself are moving; but what a representation should make explicit is a long-standing question in vision. One tradition answered that a general-purpose description of the scene should come first, built once for any task to read \citep{marr1982vision}. The other answered that perception is purposive: it exists for action and should pick out only what the behavior at hand needs \citep{gibson1979ecological}. Deep learning gave the second answer its mechanism: the task's own error, back-propagated into the representation \citep{lecun2015deep}.

Vision foundation models have since made the general-purpose tradition practical; navigation, building on them, has largely set the purposive one aside. A vision foundation model is a general-purpose description at a scale no one could have built by hand, but it makes explicit what its own objective needs and by its final layer has left out the rest, including much of the spatial structure it built along the way \citep{yosinski2014transferable,papyan2020prevalence,raghu2021vision}; most navigation policies built on vision-language models take that output through an interface fixed before the task exists \citep{zhang2025navfom,zhang2024uninavid,internvla2025}. Fine-tuning the encoder is only a partial remedy: it adapts one network behind the same output and can erode the general description \citep{kumar2022finetuning}. They follow open-ended instructions and track a person described in words, yet still misjudge free space, collide with obstacles they could describe, and lose a target that turns a corner. The field has more general-purpose visual prior than ever, yet its policies have little say over what they see.

We argue that the two answers need no longer exclude each other in embodied navigation. With vision foundation models a general-purpose description now exists, so the purposive step can be taken on top of it rather than instead of it, and the action gradient is the instrument for taking it. We call the result a \emph{self-constructed representation space}: a representation constructed by the policy's own objectives out of the internal structure of vision foundation models. We present \ours{} (RepNav), a navigation foundation model that sees and dreams in such a space. It builds the space with \mixres (Fig.~\ref{fig:representnav_overview}a), a compact three-layer transformer that attends over layers spanning the full depth of three complementary vision foundation models, semantic, geometric, and dynamic; what it makes explicit is found in training and differs across regions of one view. This is how \ours sees what it needs.

The same question arises in future prediction. Seeing tells the robot where the free space is; crossing the hall requires knowing where it will be a moment later, once the things around it have moved, and an agent that carries a model of the world can anticipate these changes before it acts \citep{sutton1991dyna,ha2018world,lecun2022path}. Latent reasoning and navigation world models do this, but in spaces defined by someone else: text, hand-designed spatial tokens, the latent space of a pretrained generative model, or pixels \citep{hao2024coconut,lin2024navcot,liu2026spannav,zuo2026fantasyvln,bar2025navigation}. A policy that builds its own space for acting has a space of its own to dream in. \ours therefore precedes every action with a latent chain-of-thought (Fig.~\ref{fig:representnav_overview}b), trained to predict the representation of the scene 0.4 to 1.6\,s ahead in that same space. Its targets are the [CLS] tokens of the future observations, a learnable token that attends alongside the spatial tokens through the three layers of \mixres and gathers each view into one vector; the dream is learned jointly with the policy as a joint-embedding predictive architecture (JEPA), with a next-representation prediction loss and a VISReg regularizer~\citep{wu2026visreg} on the targets. This is how \ours dreams what it wants: in the space through which it sees.

\ours sets a new state of the art on VLNVerse, VLN-CE R2R, EVT-Bench, and two point-goal benchmarks (Figure~\ref{fig:representnav_teaser}), including 83.80\% success on VLNVerse fine-grained (+20.05) and 85.5\% tracking rate on EVT-Bench distracted tracking (+10.1) with a single camera, and the gains carry over to a physical robot (Section~\ref{sec:experiments}). Because our claims concern the representation, we then look inside (Section~\ref{sec:analysis}) and find:
\begin{itemize}[leftmargin=*,nosep]
\item \textbf{The self-constructed space exploits the internal structure of the priors.} Without a dream, it nearly doubles the gain of concatenating the same three models' outputs, drawing on layers concatenation never sees, such as the dynamic model's layer before its last, and differently across regions of a view (Section~\ref{sec:analysis_see}).
\item \textbf{The dream pays off in the self-constructed space.} Dreaming in the space the policy sees through removes a quarter of the collisions (Section~\ref{sec:analysis_space}).
\item \textbf{One dreamed latent holds a compact world state.} Decoded after the fact, it gives the scene's depth, semantics, and 3D occupancy more accurately than pretrained encoders applied to the real future frame, though occupancy was never a training target, and the rollout gives how the robot and the target move up to 1.2\,s ahead (Section~\ref{sec:analysis_dream}).
\end{itemize}

In summary, we build a purposive representation on top of a general-purpose one and realize it in \ours with \mixres, a latent chain-of-thought trained as a JEPA in the constructed space, and a two-stage curriculum that first makes the policy act through its dream and then lets it act on what it sees and dreams together.

%% file: section/related_works.tex
\section{Related Work}\label{sec:related}

\paragraph{Foundation models as the policy's view.}
Which pretrained representation to hand a policy is a long-standing question in robot learning \citep{nair2022r3m,majumdar2023vc1,shang2024theia}. Since fine-tuning erodes the prior \citep{kumar2022finetuning}, most policies freeze the encoders and fix the interface by hand: the final layer, a fixed set of intermediate layers \citep{ranftl2021dpt,lin2025depthanything3,meng2024deepstack,qwen3vl2025}, or several encoders merged by concatenation, routing, or distillation \citep{tong2024cambrian,zong2024mova,ranzinger2024amradio,wang2026ver}. Building on attention residuals \citep{chen2026attnres}, \mixres construct the interface with the task's own objective instead, over layers spanning the full depth of several vision foundation models.

\paragraph{Predicting before acting.}
World models let an agent predict before it acts \citep{ha2018world,hafner2020dreamer}. Navigation has predicted in spaces chosen in advance, pixels, text, occupancy-grounded tokens, or the latent space of a pretrained generative model \citep{bar2025navigation,lin2024navcot,gao2025octonav,liu2026spannav,zuo2026fantasyvln}, and continuous latent reasoning thinks without words \citep{hao2024coconut,liu2026last,chen2026lastr1,liu2026lasthd}. Joint-embedding predictive architectures learn the space along with the predictor \citep{lecun2022path,assran2025vjepa2,chahe2026pijepa,zhao2026frappe} and keep it from collapsing with a moving-average target \citep{grill2020byol} or a regularizer \citep{bardes2022vicreg,balestriero2025lejepa,wu2026visreg}; LeWorldModel trains one end-to-end from pixels, adding only a regularizer \citep{maes2026leworldmodel}. \ours trains its latent chain-of-thought as one, in the space the policy sees through: the regularizer keeps that space from collapsing, and the task's own objective decides what it holds.

\paragraph{Navigation foundation models.}
Navigation foundation models unify several navigation tasks on a vision-language backbone \citep{zhang2025navfom,zhang2024uninavid,internvla2025,wang2025trackvla,zhang2026qwenrobotnav,gong2026abot,amap2026abotn1}, behind a visual interface fixed before the task exists. \ours keeps this generality and changes what lies beneath it: the space the backbone sees through, and dreams in, is constructed by the task's own objective.

%% file: section/method.tex
\section{Method}
\label{sec:method}

\ours plans in a representation space it constructs for itself (Figure~\ref{fig:representnav_overview}). Given a language instruction $L$ and RGB observations $I_{1:t}^{1:N}$ from $N\geq1$ cameras, it predicts eight waypoints $\tau=\{(x_i,y_i,\theta_i)\}_{i=1}^{8}$ in the current robot frame, each a position and a yaw, through one output shared by point-goal navigation, tracking, and vision-and-language navigation (VLN); image and video question answering (QA) is co-trained on the same Qwen3-VL backbone~\citep{qwen3vl2025} and answered autoregressively. Section~\ref{sec:mixres} constructs the space with \mixres, which attend over layers spanning the full depth of three frozen vision foundation models (\emph{see what I need}); Section~\ref{sec:representation_learning} rolls out a latent chain-of-thought that predicts the space a moment ahead before the policy acts (\emph{dream what I want}); Section~\ref{sec:training} trains both in two stages.

\IfFileExists{fig/images/representnav_pipeline.pdf}{%
\begin{figure}[t]
    \centering
    \ifdefined\includegraphics
        \includegraphics[width=\linewidth]{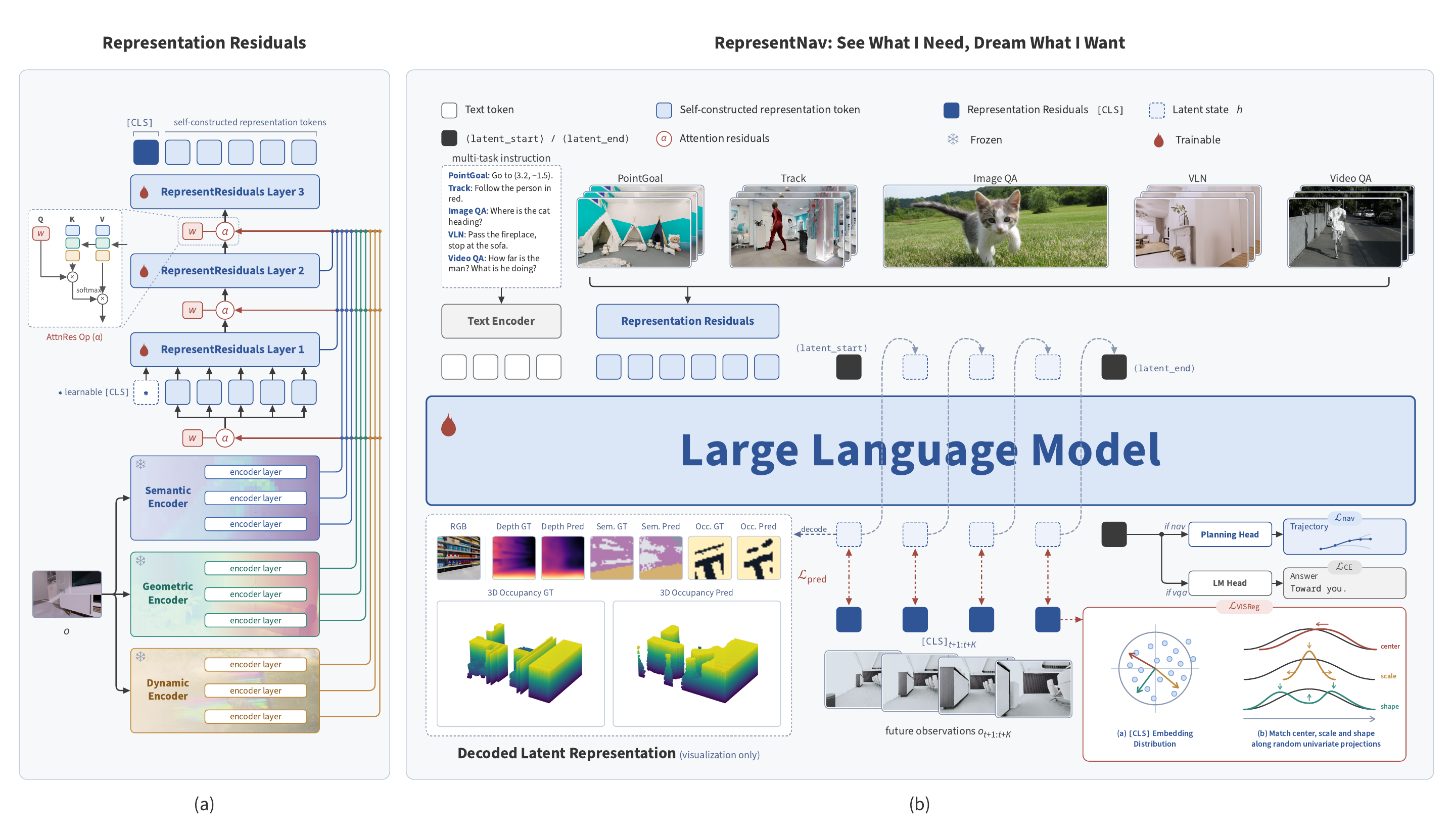}
    \else
        \pdfximage width \linewidth {fig/images/representnav_pipeline.pdf}
        \pdfrefximage\pdflastximage
    \fi
    \caption{\textbf{\ours overview.} (a) \mixres attend over layers spanning the full depth of three frozen vision foundation models and build the self-constructed representation, summarized per view by a learnable [CLS] token. (b) The policy dreams in this space: a latent chain-of-thought predicts the [CLS] tokens of future observations, regularized by VISReg, before the planning head acts; QA shares the backbone, and decoded latents are for visualization only.}
    \label{fig:representnav_overview}
\end{figure}
}{}

\subsection{\texorpdfstring{\mixres}{Representation Residuals}: See What I Need}
\label{sec:mixres}

Vision foundation models make different things explicit at different depths: the semantic, geometric, and dynamic structure that navigation needs is spread across encoders and across their layers, and the final layer keeps only part of it. \mixres attend dynamically over layers spanning the full depth of several encoders and construct from them a representation of the policy's own; since they are trained with the policy, the task's own objective decides what this space makes explicit.

For each observation $I_{t'}^v$ at time $t'$ and view $v$, DINOv3~\citep{simeoni2025dinov3} (semantic) and Depth Anything 3~\citep{lin2025depthanything3} (geometric) encode the frame and the video encoder of D4RT~\citep{zhang2025d4rt} (dynamic), in its open implementation OpenD4RT~\citep{opend4rt}, the same-view clip ending at $t'$, all three frozen; light adapters align features from layers spaced evenly over each encoder's depth to the spatial size of the backbone's native visual input (Appendix~\ref{sec:appendix_training}). \mixres are three transformer layers built on attention residuals~\citep{chen2026attnres}. At layer $\ell$ and position $p$, with $f_{\ell,0}(p)$ the incoming hidden state and $f_{\ell,j}(p)$, $j=1,\ldots,J$, the aligned encoder--layer features,
\begin{equation}
\alpha_{\ell,j}(p)=\operatorname{softmax}_j\!\big[q_\ell^\top\operatorname{RMSNorm}\big(f_{\ell,j}(p)\big)+b_\ell\mathbf{1}[j=0]\big],\qquad
\widetilde f_\ell(p)=\sum_{j=0}^{J}\alpha_{\ell,j}(p)\,f_{\ell,j}(p),
\label{eq:fusion_weights}
\end{equation}
with a learned query $q_\ell$ and a bias $b_\ell$ on the hidden state. The attended spatial tokens and a learnable [CLS] token then pass through residual self-attention and a feed-forward layer. Across the three layers the [CLS] token gathers each view into one vector, the global representation from which Section~\ref{sec:representation_learning} builds its targets. Each view enters the language model as its [CLS] token and a few spatial tokens, finer for the current view and coarser for history, alongside the backbone's own visual tokens (token counts in Appendix~\ref{sec:appendix_training}).

\subsection{Latent Chain-of-Thought: Dream What I Want}
\label{sec:representation_learning}

Anticipating the world a moment ahead gives a navigation policy foresight, but what is worth anticipating depends on the task: free space for point-goal navigation, the target's next position for tracking. Prior models choose the prediction target by hand, occupancy in SPAN-Nav~\citep{liu2026spannav} and the target's polar position in TrackVLA++~\citep{liu2025trackvlapp}. \ours predicts in the space it sees through, so the task's own objective decides what the dream holds. At inference, the rollout is a latent chain-of-thought that gives the planner this foresight before it acts; in training, it is a joint-embedding predictive architecture (JEPA)~\citep{lecun2022path,maes2026leworldmodel} that predicts representations rather than pixels.

\noindent\textbf{Targets.} At each future time $t+\Delta_k$, the shared \mixres encode the real observations into per-view [CLS] tokens $c^v_{t+\Delta_k}$, which a learnable projector $G$ maps, concatenated in front--right--back--left order, to the target
\begin{equation}
z_k=G\!\left(\operatorname{Concat}_{v\in(F,R,B,L)}c^v_{t+\Delta_k}\right),\qquad k=1,\ldots,4,
\label{eq:panoramic_representation}
\end{equation}
with $\Delta=(2,4,6,8)$, that is $0.4$ to $1.6$\,s ahead at 5\,Hz. Future observations only build targets and never enter the policy's input, and each target uses observations only up to its own time. Because \mixres and $G$ stay trainable, the targets move with the space (dimensions in Appendix~\ref{sec:appendix_training}).

\noindent\textbf{Rollout.} After the context $X$ of visual tokens, instruction, and time and view identifiers, and a latent-start token, the language model generates four continuous latents, each fed back as the next input embedding, and a projector maps them to the target space:
\begin{equation}
h_k=\operatorname{LLM}(X,h_{<k}),\qquad \widehat z_k=\operatorname{Proj}(h_k),\qquad
L_{\mathrm{wm}}=\operatorname{MSE}\big(\{\widehat z_k\}_{k=1}^{4},\{z_k\}_{k=1}^{4}\big),
\label{eq:latent_rollout}
\end{equation}
where $\operatorname{LLM}$ returns the hidden state at the last position. The rollout is free-running in training as at inference. The loss averages over target times with available observations, and gradients flow into both the prediction and the target, with no stop-gradient and no moving average, as in LeWorldModel~\citep{maes2026leworldmodel}. A latent-end token follows the rollout, and the planning head maps its hidden state to the eight waypoints.

\noindent\textbf{Keeping the space from collapsing.} Because the target comes from the same \mixres being trained, prediction alone could collapse it. We therefore apply VISReg~\citep{wu2026visreg} to the targets of each future time, gathered over the global batch into $Z\in\mathbb{R}^{N\times D}$ with mean $\mu$ and per-dimension standard deviation $\sigma$, and average it over the four times:
\begin{align}
L_{\mathrm{reg}}&=L_{\mathrm{center}}+L_{\mathrm{scale}}+L_{\mathrm{shape}},\label{eq:visreg}\\
L_{\mathrm{center}}=\tfrac{1}{D}\lVert\mu\rVert_2^2,\quad
L_{\mathrm{scale}}&=\tfrac{1}{D}\textstyle\sum_{j=1}^{D}(1-\sigma_j)^2,\quad
L_{\mathrm{shape}}=\tfrac{1}{NK}\textstyle\sum_{k=1}^{K}\big\lVert\operatorname{sort}(\widetilde Z w_k)-\mathbf{q}\big\rVert_2^2,\notag
\end{align}
where $\widetilde Z=(Z-\mu)/\operatorname{sg}(\sigma)$, $w_k$ are $K$ random unit directions, $q_i=\Phi^{-1}\big(i/(N+1)\big)$ are the quantiles of a standard Gaussian, and the stop-gradient $\operatorname{sg}$ keeps the shape term from acting on the scale. The three terms pull the targets toward zero mean, unit variance, and a Gaussian shape along every sampled direction. They fix the scale and shape of the space and nothing else; navigation and prediction decide its content.

\subsection{Navigation-Grounded Predictive Curriculum}
\label{sec:training}

Training starts from pretrained Qwen3-VL-4B, with \mixres, the [CLS] embedding, the planning head, and both projectors newly initialized, and mixes point-goal, tracking, VLN, and QA data under
\begin{equation}
L=\lambda_{\mathrm{nav}}L_{\mathrm{nav}}+\lambda_{\mathrm{wm}}L_{\mathrm{wm}}+\lambda_{\mathrm{reg}}L_{\mathrm{reg}}+\lambda_{\mathrm{qa}}L_{\mathrm{qa}},
\label{eq:stage_objective}
\end{equation}
where $L_{\mathrm{nav}}$ is the MSE between normalized predicted and expert trajectories and $L_{\mathrm{qa}}$ the answer-token cross-entropy. \textbf{Stage I} makes the policy act through its dream: following the attention curriculum of LaST-VLA~\citep{luo2026lastvla}, the instruction and the observations are isolated from each other, the latents attend to both, and the action position attends only to the instruction and the latents; the loss weights favor prediction. \textbf{Stage II} continues from Stage I with full causal attention, so the policy acts on what it sees and what it dreams together, and the weights shift to the trajectory. The training samples are split into two disjoint halves, with one half used in Stage I and the other in Stage II. The three vision foundation models and the backbone's visual encoder stay frozen throughout; loss weights, trainable modules, and the QA path are in Appendix~\ref{sec:appendix_training}.

\subsection{Training Data}
\label{sec:data}

The model is trained on about 5.08M samples (Appendix Figure~\ref{fig:data_composition}): 4.69M for navigation, namely 2.52M VLN samples from R2R~\citep{anderson2018vision} and RxR~\citep{ku2020room}, 0.50M fine- and coarse-grained samples from VLNVerse~\citep{lin2025vlnverse}, 0.67M tracking samples~\citep{wang2025trackvla}, and 1.0M point-goal samples including SAGE-3D~\citep{miao2025sage3d}; and 0.39M QA samples from VQA v2~\citep{goyal2017vqav2} and CLEVR~\citep{johnson2017clevr}. For VLNVerse we follow the training setup of TANGO~\citep{li2026tango}. All navigation samples share one format: an observation history from one or more cameras, a task instruction or goal, and future waypoints as supervision. The two stages train on disjoint halves of the samples, each drawn evenly from every source and task.

%% file: section/experiments.tex
\section{Experiments}\label{sec:experiments}
\subsection{Experimental Setup}
We evaluate \ours on vision-and-language navigation, active visual tracking, and point-goal navigation across different environments and observation settings. Section~\ref{sec:analysis} presents component ablations and representation analyses.

\paragraph{Benchmarks and metrics.}
We evaluate on five benchmarks: VLN-CE (R2R and RxR Val-Unseen)~\citep{krantz2020beyond,anderson2018vision,ku2020room} and VLNVerse (Fine- and Coarse-grained, Val-Unseen and Full Test, 3\,m success threshold)~\citep{lin2025vlnverse} for vision-and-language navigation; EVT-Bench (single-target and distracted tracking)~\citep{wang2025trackvla} for active tracking; and, for point-goal navigation, the InternVLA-N1 protocol~\citep{internvla2025} on InternScenes Home and Commercial~\citep{wang2025internscenes} and indoor ABotN-PointBench~\citep{amap2026abotn1}, whose success requires collision-free arrival. Metrics follow each benchmark: NE (m), OSR, SR, SPL, and nDTW for VLN; TR, SR, and CR for tracking; SR and SPL for point-goal. Rates are on a 0--100 scale; dashes denote unreported results.

\paragraph{Implementation Details.}
We train with Muon (learning rate $10^{-3}$) for matrix parameters and AdamW ($10^{-5}$) for the rest, a global batch of 128, and 38,270 steps in total on 16 GPUs. Token counts and the loss weights of each stage are in Appendix~\ref{sec:appendix_training}.

\subsection{Benchmark Results}
\label{sec:benchmark-results}

\begin{table}[p]
\centering
\caption{\textbf{Benchmark results.} Baselines are from the cited works; \textbf{bold}: best per column. (a) EVT-Bench single-target (STT) and distracted (DT) tracking \citep{liu2025trackvlapp,zhang2026qwenrobotnav,gong2026abot,amap2026abotn1}; best per observation setting. (b) InternVLA-N1 point-goal \citep{liu2026spannav,xu2025mmnav}. (c) Indoor ABotN-PointBench, collision-free success \citep{amap2026abotn1}. (d) VLN-CE Val-Unseen \citep{liu2026spannav,wei2026dualvln,internvla2025,amap2026abotn1,zhang2025navfom,zhang2026qwenrobotnav}; best per observation setting; $^*$: waypoint-predictor-based; Qwen-RobotNav at 4B, the scale of our backbone. (e) VLNVerse, 3\,m threshold, Full Test \citep{zhang2026qwenrobotnav} and Val-Unseen \citep{lin2025vlnverse,li2026tango}; best per split.}
\label{tab:bench}
\fontsize{9}{10.8}\selectfont\setlength{\tabcolsep}{4pt}
\subtab{(a) EVT-Bench}
\begin{tabular*}{\linewidth}{@{\extracolsep{\fill}}l rrr rrr@{}}
\toprule
& \multicolumn{3}{c}{STT} & \multicolumn{3}{c}{DT} \\
\cmidrule(lr){2-4}\cmidrule(l){5-7}
Method & TR$\uparrow$ & SR$\uparrow$ & CR$\downarrow$ & TR$\uparrow$ & SR$\uparrow$ & CR$\downarrow$ \\
\midrule
\multicolumn{7}{l}{\emph{Monocular}} \\
Uni-NaVid & 39.5 & 25.7 & 41.90 & 27.4 & 11.3 & 43.50 \\
TrackVLA & 78.6 & 85.1 & \textbf{1.65} & 63.2 & 57.6 & 5.80 \\
NavFoM & 80.5 & 85.0 & -- & 68.2 & 61.4 & -- \\
TrackVLA++ & 81.0 & 86.0 & 2.10 & 68.8 & 66.5 & \textbf{4.71} \\
ABot-N0 & 87.6 & 86.9 & 8.54 & 75.4 & 66.7 & 11.60 \\
Qwen-RobotNav-4B & 90.0 & 77.4 & 6.40 & -- & -- & -- \\
Qwen-RobotNav-8B & 89.7 & 78.6 & 5.70 & -- & -- & -- \\
\textcolor{pkured}{\textbf{RepNav}} & \textbf{92.7} & \textbf{87.1} & 4.63 & \textbf{85.5} & \textbf{68.3} & 11.89 \\
\midrule
\multicolumn{7}{l}{\emph{Multi-view}} \\
NavFoM & 80.7 & 88.4 & -- & 67.9 & 62.0 & -- \\
TrackVLA++ & 82.7 & \textbf{90.9} & \textbf{1.50} & 73.7 & 74.0 & \textbf{3.51} \\
ABot-N1 & 89.8 & 90.1 & 4.27 & 84.4 & 67.4 & 17.90 \\
\textcolor{pkured}{\textbf{RepNav}} & \textbf{92.9} & 90.0 & 2.85 & \textbf{86.8} & \textbf{75.6} & 8.11 \\
\bottomrule
\end{tabular*}

\par\vspace{9pt}
\subtab{(b) InternVLA-N1 point-goal}
\begin{tabular*}{\linewidth}{@{\extracolsep{\fill}}l rr rr rr@{}}
\toprule
& \multicolumn{2}{c}{Home} & \multicolumn{2}{c}{Commercial} & \multicolumn{2}{c}{Mean} \\
\cmidrule(lr){2-3}\cmidrule(lr){4-5}\cmidrule(l){6-7}
Method & SR$\uparrow$ & SPL$\uparrow$ & SR$\uparrow$ & SPL$\uparrow$ & SR$\uparrow$ & SPL$\uparrow$ \\
\midrule
InternVLA-N1 & 60.0 & 55.6 & 71.4 & 68.2 & 65.7 & 61.9 \\
NavDP & 60.3 & 54.7 & 74.1 & 70.5 & 67.2 & 62.6 \\
SIDP & 63.2 & 56.5 & 81.2 & 73.4 & 72.2 & 65.0 \\
MM-Nav (mono.) & 79.9 & 77.6 & 86.7 & 84.7 & 83.3 & 81.2 \\
MM-Nav (multi.) & 86.3 & 81.1 & 89.9 & 85.5 & 88.1 & 83.3 \\
SPAN-Nav & 90.9 & \textbf{85.7} & 91.0 & 88.1 & 91.0 & \textbf{86.9} \\
\textcolor{pkured}{\textbf{RepNav}} & \textbf{95.5} & 83.4 & \textbf{94.5} & \textbf{89.9} & \textbf{95.0} & 86.7 \\
\bottomrule
\end{tabular*}

\par\vspace{9pt}
\subtab{(c) ABotN-PointBench}
\begin{tabular*}{\linewidth}{@{\extracolsep{\fill}}l rrr rrr@{}}
\toprule
& \multicolumn{3}{c}{SR$\uparrow$} & \multicolumn{3}{c}{SPL$\uparrow$} \\
\cmidrule(lr){2-4}\cmidrule(l){5-7}
Method & All & Low & High & All & Low & High \\
\midrule
GNM & 26.7 & 30.8 & 22.5 & 26.6 & 30.6 & 22.5 \\
ViNT & 27.9 & 32.5 & 23.3 & 27.9 & 32.4 & 23.3 \\
NoMaD & 20.0 & 20.8 & 19.2 & 19.6 & 20.5 & 18.6 \\
CityWalker & 21.7 & 23.3 & 20.0 & 21.6 & 23.3 & 19.9 \\
SocialNav & 42.5 & 46.7 & 38.3 & 42.5 & 46.7 & 38.3 \\
ABot-N0 & 89.6 & 91.7 & 87.5 & 85.6 & 87.5 & 83.7 \\
ABot-N1 & 95.4 & 95.0 & \textbf{95.8} & 93.7 & 93.3 & \textbf{94.2} \\
\textcolor{pkured}{\textbf{RepNav}} & \textbf{97.5} & \textbf{100} & 95.0 & \textbf{95.8} & \textbf{97.6} & 93.9 \\
\bottomrule
\end{tabular*}

\end{table}

\begin{table}[p]
\ContinuedFloat
\centering
\caption[]{\textbf{Benchmark results (continued).} Vision-and-language navigation. Panel descriptions and evaluation conventions are given on the preceding page.}
\fontsize{9}{10.8}\selectfont\setlength{\tabcolsep}{4pt}
\subtab{(d) VLN-CE Val-Unseen}
\begin{tabular*}{\linewidth}{@{\extracolsep{\fill}}l rrrr rrrr@{}}
\toprule
& \multicolumn{4}{c}{R2R} & \multicolumn{4}{c}{RxR} \\
\cmidrule(lr){2-5}\cmidrule(l){6-9}
Method & NE$\downarrow$ & OSR$\uparrow$ & SR$\uparrow$ & SPL$\uparrow$ & NE$\downarrow$ & SR$\uparrow$ & SPL$\uparrow$ & nDTW$\uparrow$ \\
\midrule
\multicolumn{9}{l}{\emph{Monocular}} \\
NaVid & 5.72 & 49.2 & 41.9 & 36.5 & 5.72 & 45.7 & 38.2 & -- \\
Uni-NaVid & 5.58 & 53.3 & 47.0 & 42.7 & 6.24 & 48.7 & 40.9 & -- \\
NaVILA & 5.22 & 62.5 & 54.0 & 49.0 & 6.77 & 49.3 & 44.0 & 58.8 \\
StreamVLN & 4.98 & 64.2 & 56.9 & 51.9 & 6.22 & 52.9 & 46.0 & 61.9 \\
InternVLA-N1 & 4.83 & 63.3 & 58.2 & 54.0 & 5.91 & 53.5 & 46.1 & 65.3 \\
Qwen-VLA & 5.10 & 69.0 & 57.3 & 51.2 & 5.80 & 59.6 & 47.8 & -- \\
DualVLN & 4.05 & 70.7 & 64.3 & 58.5 & 4.58 & 61.4 & 51.8 & \textbf{70.0} \\
CorrectNav & 4.24 & 67.5 & 65.1 & \textbf{62.3} & \textbf{4.09} & \textbf{69.3} & \textbf{63.3} & -- \\
NavForesee & \textbf{3.94} & \textbf{78.4} & \textbf{66.2} & 59.7 & 4.20 & 66.3 & 53.2 & -- \\
\midrule
\multicolumn{9}{l}{\emph{Multi-view}} \\
CMA$^*$ & 6.20 & 52.0 & 41.0 & 36.0 & 8.76 & 26.5 & 22.1 & 47.0 \\
VLN-BERT$^*$ & 5.74 & 53.0 & 44.0 & 39.0 & 8.98 & 27.0 & 22.6 & 46.7 \\
AO-Planner & 5.55 & 59.0 & 47.0 & 33.0 & 7.06 & 43.3 & 30.5 & 50.1 \\
Reborn$^*$ & 5.40 & 57.0 & 50.0 & 46.0 & 5.98 & 48.6 & 42.0 & 63.3 \\
ETPNav$^*$ & 4.71 & 65.0 & 57.0 & 49.0 & 5.64 & 54.7 & 44.8 & 61.9 \\
HNR$^*$ & 4.42 & 67.0 & 61.0 & 51.0 & 5.50 & 56.3 & 46.7 & 63.5 \\
NavFoM & 4.61 & 72.1 & 61.7 & 55.3 & 4.74 & 64.4 & 56.2 & 65.8 \\
SPAN-Nav & 4.07 & 75.3 & 66.3 & 59.3 & 4.20 & 69.7 & 60.1 & 67.9 \\
ABot-N0 & 3.78 & 70.8 & 66.4 & 63.9 & 3.83 & 69.3 & 60.0 & -- \\
Qwen-RobotNav-4B & 3.80 & \textbf{77.2} & 69.5 & 63.6 & \textbf{3.80} & \textbf{75.2} & \textbf{65.0} & \textbf{71.9} \\
\textcolor{pkured}{\textbf{RepNav}} & \textbf{3.75} & 73.8 & \textbf{71.1} & \textbf{66.6} & 3.85 & 73.0 & 64.2 & \textbf{71.9} \\
\bottomrule
\end{tabular*}

\par\vspace{9pt}
\subtab{(e) VLNVerse}
\begin{tabular*}{\linewidth}{@{\extracolsep{\fill}}l rrrr rrrr@{}}
\toprule
& \multicolumn{4}{c}{Fine-grained} & \multicolumn{4}{c}{Coarse-grained} \\
\cmidrule(lr){2-5}\cmidrule(l){6-9}
Method & NE$\downarrow$ & OSR$\uparrow$ & SR$\uparrow$ & SPL$\uparrow$ & NE$\downarrow$ & OSR$\uparrow$ & SR$\uparrow$ & SPL$\uparrow$ \\
\midrule
\multicolumn{9}{l}{\emph{Full Test}} \\
InternVLA-N1 & 5.68 & 38.69 & 28.95 & 25.00 & 5.71 & 23.32 & 17.51 & 16.54 \\
Uni-NaVid & 4.11 & 67.02 & 45.74 & 26.91 & 5.05 & 42.70 & 29.68 & 13.47 \\
NavFoM & 4.13 & 69.68 & 51.59 & 32.40 & 4.93 & 45.93 & 38.02 & 23.15 \\
Qwen-RobotNav-4B & 3.05 & 70.36 & 62.61 & 56.22 & 4.40 & 51.03 & 41.25 & 37.37 \\
Qwen-RobotNav-8B & 3.00 & 72.81 & 63.75 & 57.93 & 4.08 & 55.97 & 46.59 & 41.54 \\
\textcolor{pkured}{\textbf{RepNav}} & \textbf{1.60} & \textbf{89.00} & \textbf{83.80} & \textbf{80.90} & \textbf{3.30} & \textbf{69.10} & \textbf{59.00} & \textbf{57.40} \\
\midrule
\multicolumn{9}{l}{\emph{Val-Unseen}} \\
CMA & 5.16 & 62.79 & 31.15 & 27.92 & -- & -- & 35.52 & 32.48 \\
Seq2Seq & 4.36 & 49.58 & 35.03 & 33.37 & -- & -- & 33.09 & 31.00 \\
HNR & -- & -- & 32.95 & 29.56 & -- & -- & 40.92 & 37.27 \\
RDP & 3.75 & 71.93 & 48.60 & 42.72 & -- & -- & 40.85 & 35.13 \\
GAMA & -- & -- & 38.69 & 34.38 & -- & -- & 48.30 & 41.50 \\
InternVLA-N1 & 4.09 & 64.13 & 45.56 & 34.98 & -- & -- & -- & -- \\
Uni-NaVid & 3.97 & 62.50 & 45.00 & 39.42 & -- & -- & -- & -- \\
TANGO & 3.72 & 71.07 & 52.89 & 40.18 & -- & -- & -- & -- \\
\textcolor{pkured}{\textbf{RepNav}} & \textbf{1.42} & \textbf{92.73} & \textbf{89.21} & \textbf{86.50} & -- & -- & \textbf{64.67} & \textbf{62.86} \\
\bottomrule
\end{tabular*}

\end{table}

\paragraph{Active visual tracking.}
\ours has the highest tracking rate in both observation settings of EVT-Bench (Table~\ref{tab:bench}(a)). With a single camera it reaches 85.5 on distracted tracking, 10.1 points above the best monocular baseline and above every multi-view baseline; with multiple views it reaches 86.8. TrackVLA and TrackVLA++ keep lower collision rates.

\paragraph{Point-goal navigation.}
On the InternVLA-N1 benchmark, \ours raises mean success to 95.0\%, 4.0 points above SPAN-Nav (Table~\ref{tab:bench}(b)). On ABotN-PointBench, where an episode counts only if the robot arrives without a collision, it reaches 97.5\% success and 95.8 SPL, 2.1 points above ABot-N1 on both (Table~\ref{tab:bench}(c)).

\paragraph{Vision-and-language navigation.}
On the VLNVerse test split, \ours reaches 83.80\% success with fine-grained instructions and 59.00\% with coarse-grained ones, 20.05 and 12.41 points above Qwen-RobotNav-8B; it also leads every reported metric on Val-Unseen (Table~\ref{tab:bench}(e)). Among multi-view methods on VLN-CE, it has the best R2R success, 71.1\%, SPL, and navigation error, and ties Qwen-RobotNav-4B on RxR nDTW, 71.9, while Qwen-RobotNav-4B leads the other RxR metrics (Table~\ref{tab:bench}(d)).

\subsection{Real-World Results}

\begin{figure}[t]
    \centering
    \includegraphics[width=\linewidth]{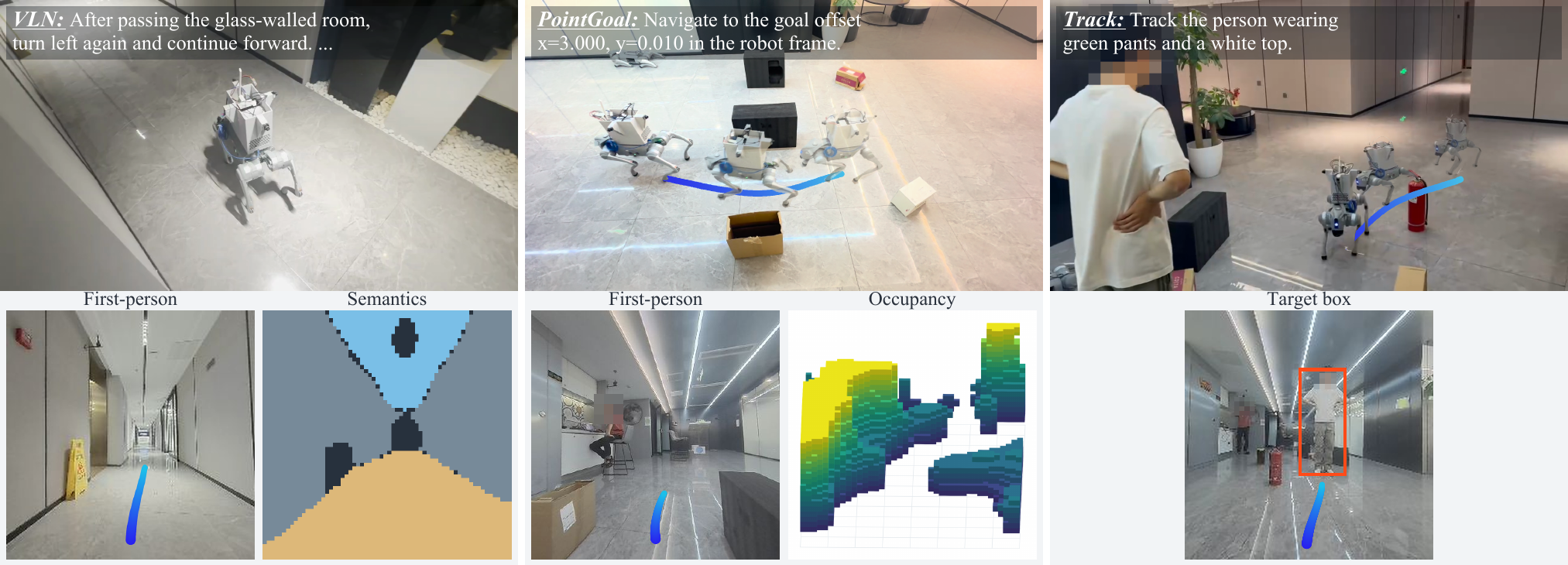}
    \caption{\textbf{Real-world navigation and representation readouts.}
    \ours{} performs point-goal navigation, person following, and vision-and-language navigation in real environments. Insets show occupancy, target-box, and semantic readouts.}
    \label{fig:real_world}
\end{figure}

We further evaluate \ours{} in unseen real-world environments across three navigation tasks.
As shown in Figure~\ref{fig:real_world}, the model avoids obstacles, follows a designated person, and navigates according to language instructions, demonstrating strong cross-environment generalization.
Occupancy decoded from the dream latents preserves spatial information relevant to path planning and obstacle avoidance, while bounding-box and semantic readouts reveal the model's perception of targets and scene layout.
We also report quantitative real-world evaluations on two robotic platforms, a Go2 quadruped and a Bumi humanoid, in Appendix~\ref{sec:appendix_qualitative}.

\FloatBarrier

%% file: section/analysis.tex
\input{section/analysis_20260920_final_en}

%% file: section/analysis_20260920_final_en.tex
\section{Ablation Study and Analysis}\label{sec:analysis}

This section examines where these gains come from, on the task of Section~\ref{sec:intro}: a robot crossing a cluttered hall, here the InternVLA-N1 PointGoal protocol~\citep{internvla2025} on 4,000 trajectories. Success is near its ceiling there, so the differences show mainly in collisions.

\begin{table}[!htbp]
\centering\footnotesize\setlength{\tabcolsep}{4pt}
\caption{\textbf{What the policy sees and where it dreams.} Point-goal ablations; no row in (a) dreams. General description: semantic (DINOv3), geometric (DA3), and dynamic (OpenD4RT) vision foundation models. W/o supervision: the latents are rolled out with no loss on them.}
\label{tab:ablation}\label{tab:see_ladder}\label{tab:dream_space}
\begin{tabular*}{\linewidth}{@{\extracolsep{\fill}}ccc l l rrr@{}}
\toprule
\multicolumn{3}{c}{General description} & & & & & \\
\cmidrule(lr){1-3}
Sem. & Geo. & Dyn. & Interface & Dream target & SR (\%)$\uparrow$ & SPL$\uparrow$ & CR (\%)$\downarrow$ \\
\midrule
\multicolumn{8}{l}{\emph{(a) What the policy sees (no dream)}} \\
-- & -- & -- & native ViT & none & 79.65 & 73.43 & 40.80 \\
-- & $\checkmark$ & -- & concat. & none & 83.25 & 76.08 & 35.40 \\
$\checkmark$ & $\checkmark$ & -- & concat. & none & 86.70 & 79.57 & 30.90 \\
$\checkmark$ & $\checkmark$ & $\checkmark$ & concat. & none & 87.30 & 80.20 & 28.70 \\
$\checkmark$ & $\checkmark$ & -- & \mixres & none & 89.80 & 82.20 & 24.30 \\
$\checkmark$ & $\checkmark$ & $\checkmark$ & \mixres & none & 93.45 & 84.28 & 18.05 \\
\midrule
\multicolumn{8}{l}{\emph{(b) What the dream predicts}} \\
$\checkmark$ & $\checkmark$ & $\checkmark$ & \mixres & w/o supervision & 93.35 & 84.34 & 15.90 \\
$\checkmark$ & $\checkmark$ & $\checkmark$ & \mixres & native, pooled & 92.75 & 82.69 & 22.00 \\
\midrule
\multicolumn{8}{l}{\emph{RepNav (ours)}} \\
$\checkmark$ & $\checkmark$ & $\checkmark$ & \mixres & self-constructed & \textbf{95.00} & \textbf{86.70} & \textbf{13.45} \\
\bottomrule
\end{tabular*}
\end{table}

\subsection{The Self-Constructed Space Exploits the Internal Structure of the Priors}
\label{sec:analysis_see}

\begin{figure}[!ht]
\centering
\includegraphics[height=1.5in,trim=0 0 278pt 0,clip]{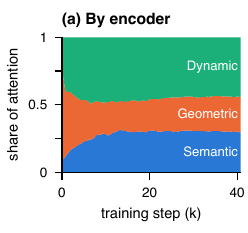}\hfill
\begin{minipage}[b]{3.56in}\raggedright
{\sffamily\bfseries\fontsize{7.5}{9}\selectfont (b) By region}\par\vspace{1pt}
\includegraphics[height=1.42in]{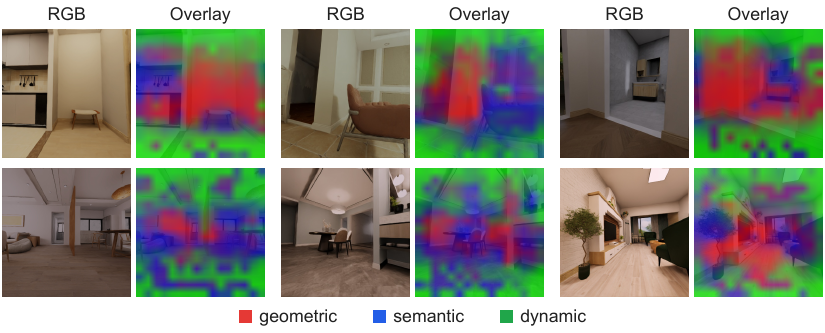}
\end{minipage}
\caption{\textbf{What the constructed space draws on.} (a) Attention shares of \mixres by encoder during training; the running state, initialized from the geometric encoder, counts as geometric. (b) Where \mixres in the final model prefer each encoder, in six selected VLNVerse views; colors show contrast-enhanced relative preferences, not absolute weights. Details and shares by depth in Appendix~\ref{sec:appendix_see}.}
\label{fig:alpha}
\end{figure}

\textbf{The interface shapes what a policy gets from foundation models.} Table~\ref{tab:ablation}(a) changes only the space the policy sees through, and no row dreams. Three vision foundation models concatenated with the backbone's features lift success from 79.65\% to 87.30\% and cut collisions from 40.80\% to 28.70\%; the same models through a space the policy constructs reach 93.45\% and 18.05\%, nearly twice the gain. The dynamic model adds 0.60 points through the fixed interface but 3.65 through the constructed space, whose attention suggests why (Figure~\ref{fig:alpha}; Appendix Figure~\ref{fig:alpha_full}): it takes 67\% of the dynamic model's share from L30 and only 30\% from the final layer, the one a fixed interface receives, and within one view it leans on the geometric encoder for walls, the semantic one for furniture, and the dynamic one for floor and ceiling. For navigation, it pays to let the policy construct the space it sees through. The same question arises for the dream: whose space should it predict in?

\subsection{The Dream Pays Off in the Self-Constructed Space}
\label{sec:analysis_space}

\textbf{The dream does best in the space the policy sees through.} Table~\ref{tab:ablation}(b) starts from the last row of (a) and changes only what the policy dreams. Without a dream it collides in 18.05\% of episodes, and rolling out the same latents without supervision lowers this to 15.90\%. A target in a space defined by someone else, the backbone ViT's pooled features, raises it to 22.00\%, while one in the self-constructed space lowers it to 13.45\%, a quarter fewer than without a dream. Its targets come from the same trainable \mixres, and VISReg keeps them from collapsing; without it, collisions rise to 16.40\% (Appendix~\ref{sec:appendix_space}). This raises the natural question: what does a dream in this space hold?

\subsection{One Dreamed Latent Holds a Compact World State}
\label{sec:analysis_dream}

\begin{figure}[!ht]
\centering
\includegraphics[width=\linewidth]{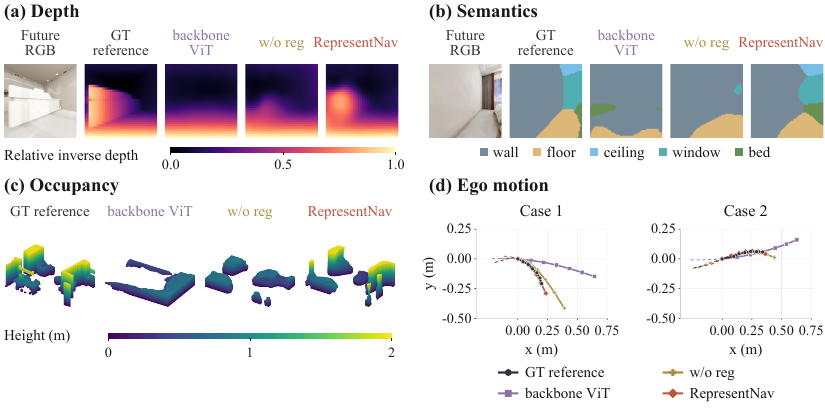}
\par\vspace{2pt}
{\footnotesize\setlength{\tabcolsep}{4pt}
\begin{tabular}{lccccc}
\toprule
& Depth & Occupancy & Semantic & Pixel & Ego motion \\
Representation & MAE$\downarrow$ & IoU (\%)$\uparrow$ & mIoU (\%)$\uparrow$ & acc. (\%)$\uparrow$ & ADE (m)$\downarrow$ \\
\midrule
Backbone ViT on real future & 0.16257 & 25.47 & 16.05 & 57.32 & 0.17932 \\
DA3 on real future & 0.14472 & 26.14 & 15.25 & 59.99 & 0.14095 \\
DINOv3 on real future & 0.15206 & 25.85 & 16.84 & 58.72 & 0.14209 \\
openD4RT on real future & 0.10253 & 29.79 & 20.34 & 66.67 & 0.04713 \\
\mixres{} [CLS] on real future & \textbf{0.07520} & \textbf{35.95} & \textbf{24.82} & \textbf{73.45} & \textbf{0.02962} \\
\midrule
w/o reg.\ (predicted) & 0.10958 & 30.99 & 18.83 & 66.36 & 0.04583 \\
RepNav (predicted) & 0.09131 & 34.81 & 23.38 & 71.35 & 0.04448 \\
\bottomrule
\end{tabular}}
\caption{\textbf{What one dreamed latent holds.} Post-hoc readouts from the $+0.4$\,s latent ((a) depth, (b) semantics, (c) 3D occupancy) and all four latents ((d) ego motion), on 1,000 test observations. Encoder features and the [CLS] tokens of \mixres from real future images are reference inputs, not predicted latents; full results are in Table~\ref{tab:readouts_full}. Examples were selected to show the model's strengths; more in Appendix Figure~\ref{fig:representnav_readouts_full}.}
\label{fig:representnav_readouts}\label{fig:depth_readouts}\label{fig:occupancy_readouts}\label{fig:semantic_readouts}\label{fig:self_motion_readouts}
\end{figure}

\textbf{One dreamed latent holds a decodable world state.} From the latent for $+0.4$\,s, decoders recover free space, surface depth, and object layout more accurately than from the pooled features of any of the four pretrained encoders applied to the real future frame, which the latent never sees (Figure~\ref{fig:representnav_readouts}). Occupancy stands out, 34.81\% IoU against at most 29.79\%, although no loss supervises it and no encoder outputs it. Across the four latents, decoders also recover the robot's displacement up to $+1.2$\,s and, on tracking, where the target is heading (Appendix~\ref{sec:track_readouts}). One latent holds the state and the rollout its change: the dream emerges as a compact latent world model.

%% file: section/conclusion.tex
\FloatBarrier
\section{Conclusion and Limitations}\label{sec:conclusion}

We asked what visual representation a navigation policy should act on, and answered with a self-constructed representation space: a purposive representation built on top of the general-purpose description of vision foundation models. \ours constructs this space with \mixres and dreams in it with a latent chain-of-thought; it sets a new state of the art across vision-and-language navigation, tracking, and point-goal navigation, carries over to a physical robot, and holds in a single dreamed latent a compact world state.

Nothing in this construction is specific to navigation: the priors are general-purpose, and what the space makes explicit is decided by whatever objective the policy is trained on. Yet we have validated it only on navigation; whether it carries over to manipulation, where latent reasoning has already proven useful~\citep{chen2026lastr1,liu2026lasthd}, dexterous hands, and humanoid whole-body control remains untested.

%% file: section/appendix.tex
\section{Method Details}\label{sec:appendix_training}

\subsection{Architecture}\label{sec:appendix_arch}

\paragraph{Encoders and adapters.} For each navigation observation at time $t'$ and view $v$, DINOv3 and DA3 process the single frame $I_{t'}^v$, while OpenD4RT processes the same-view four-frame clip $I_{t'-3:t'}^v$. Features from several layers of each encoder are aligned to a $16\times16$ grid and a common channel dimension by independent adapters, followed by source, layer, and position embeddings.

\paragraph{Token readouts.} Current and historical observations share \mixres but use different spatial readouts. For each historical view, spatial features are average-pooled over a $2\times4$ partition and projected into eight coarse tokens. For each current view, nonoverlapping $2\times2$ neighborhoods are concatenated along channels and projected into an $8\times8$ grid of 64 fine tokens. The normalized [CLS] of each view is a 1024-dimensional global representation; an independent MLP maps it to the language-model dimension for input alongside the spatial tokens. Each current view therefore contributes [CLS] plus 64 fine tokens, and each historical view [CLS] plus eight coarse tokens.

\paragraph{Targets and projectors.} Each [CLS] has 1024 dimensions; the front--right--back--left concatenation yields 4096 dimensions, which the target projector $G$ maps to 256, with missing views filled with zeros. The prediction head $\operatorname{Proj}$ maps each 2,560-dimensional hidden state to the same 256 dimensions.

\paragraph{Regularizer computation.} For each future time, target representations are gathered across the global batch and VISReg is computed separately, then averaged across times.

\subsection{Training}\label{sec:appendix_stages}

\paragraph{Stage I: learning navigation-relevant future representations.} The language backbone starts from pretrained Qwen3-VL-4B weights; \mixres, the [CLS] embedding, the planner, and the target and prediction projectors $G$ and $\operatorname{Proj}$ are newly initialized. The model uses the full visual input, performs four-step latent CoT, and constructs future targets online. Following the attention curriculum of LaST-VLA~\citep{luo2026lastvla}, we isolate the visual and language prefixes in the navigation sequence's causal mask. Latents retain access to both, while the action position attends only to language and generated latents. We set $(\lambda_{\mathrm{nav}},\lambda_{\mathrm{wm}},\lambda_{\mathrm{reg}},\lambda_{\mathrm{qa}})=(0.01,0.1,0.9,1)$, emphasizing future representation learning while retaining trajectory supervision; the QA weight of 1 in both stages offsets QA samples making up only a tenth of the training mixture.

\paragraph{Stage II: planning from observations and predictions.} We inherit the complete Stage-I model and optimization state, retain four-step latent CoT, and restore standard causal attention so that the action position can jointly read visual observations, instructions, and predicted latents. Loss weights become $(\lambda_{\mathrm{nav}},\lambda_{\mathrm{wm}},\lambda_{\mathrm{reg}},\lambda_{\mathrm{qa}})=(1,0.001,0.009,1)$, emphasizing trajectory learning while retaining future prediction and target-space regularization.

\paragraph{Trainable modules and QA.} Both stages update the language backbone, \mixres and the learnable [CLS] embedding, the planner, and both projectors. The three source visual encoders and the Qwen3-VL visual encoder remain frozen. QA samples use the backbone's visual input and features from DA3 and DINOv3 to generate answers autoregressively, without navigation latent rollout or future-target construction.

\paragraph{Data split.} The sample manifest is split once, with a fixed seed and stratified by source and task, into the two halves used by the two stages. The halves are disjoint in samples but not in episodes: the split is by sample, so one episode can contribute samples to both.

\begin{figure}[htbp]
\centering
\includegraphics[width=0.6\linewidth]{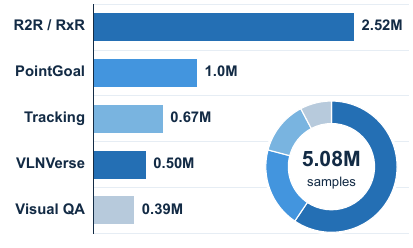}
\caption{\textbf{Training data composition} (Section~\ref{sec:data}).}
\label{fig:data_composition}
\end{figure}

\FloatBarrier
\subsection{The Two-Stage Curriculum}\label{sec:appendix_reliance}

Table~\ref{tab:curriculum} compares the curriculum of Section~\ref{sec:training} with a single stage that opens attention from the start.

\begin{table}[htbp]
\centering\small
\caption{\textbf{Effect of the two-stage curriculum.} Same model and data; the single-stage variant opens attention from the start instead of first restricting the action position to the latent chain-of-thought.}
\label{tab:curriculum}
\begin{tabular}{cc rrr}
\toprule
Restricted stage & Open stage & SR (\%)\,$\uparrow$ & SPL\,$\uparrow$ & CR (\%)\,$\downarrow$ \\
\midrule
-- & $\checkmark$ & 92.40 & 83.68 & 15.15 \\
$\checkmark$ & $\checkmark$ & \textbf{95.00} & \textbf{86.70} & \textbf{13.45} \\
\bottomrule
\end{tabular}
\end{table}

\begin{figure}[htbp]
\centering
\includegraphics[width=\linewidth]{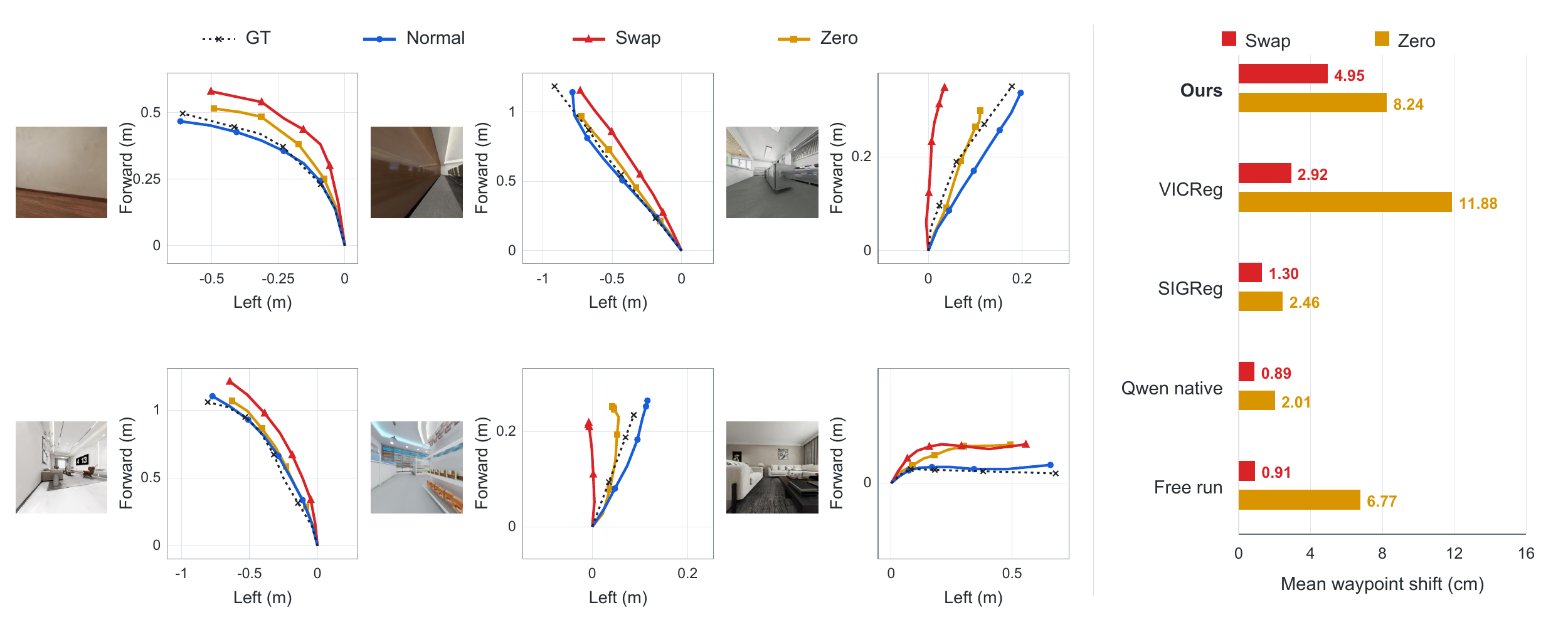}
\caption{\textbf{Latent-input interventions and action sensitivity.} Left: GT and normal, cross-scene-swap, and zero-input predictions from our two-stage point-goal policy; axes are left and forward (meters), with equal XY scales.  Right: mean XY change from each policy's normal prediction over eight waypoints on the same 200 observations. All five models use their own predicted latents; Qwen native dreams in the backbone ViT's pooled features and Free run rolls out the same latents with no loss on them (w/o supervision in Table~\ref{tab:ablation}(b)), while VICReg and SIGReg replace VISReg (Table~\ref{tab:regularizer}). Only the four recurrent inputs are swapped within each checkpoint or zeroed; other inputs stay fixed, with no future images. These open-loop diagnostics on navigation-training observations measure sensitivity, not navigation performance.}
\label{fig:reliance}
\end{figure}

\paragraph{The policy acts through its dream.} The two-stage policy relies more on the latent chain-of-thought despite similar occupancy readouts: 34.81\% IoU versus 34.57\% for the single-stage model (Tables~\ref{tab:readouts_full} and~\ref{tab:reliance}). We intervene on the latent inputs with full causal attention in both models, keeping images, history, instructions, and the action-readout input unchanged. Cross-scene swaps shift the two-stage model's predicted trajectory by 4.95\,cm on average, versus 1.38\,cm for the single-stage model; zeroing produces shifts of 8.24\,cm and 2.68\,cm (Table~\ref{tab:reliance}). Two-stage shifts are larger in 183 and 182 of the 200 observations, respectively, rather than being concentrated in a few cases. The latent chain-of-thought remains involved in action generation after full visual context is restored; disrupting its inputs also increases the two-stage model's ADE against the reference trajectory from 4.38\,cm to 7.06\,cm and 9.76\,cm. Figure~\ref{fig:reliance} shows examples and the same shifts for other dream targets and regularizers.

\begin{table}[htbp]
\centering\small
\caption{\textbf{The two-stage policy responds more strongly to changes in its latent inputs.} Each model and condition uses 200 observations.}
\label{tab:reliance}
\setlength{\tabcolsep}{5pt}
\begin{tabular}{llrrr}
\toprule
Training & Inference condition & Affected (\%) & Shift (cm) & GT ADE (cm) \\
\midrule
Single-stage & Normal & 0.0 & 0.00 & 3.93 \\
Single-stage & Cross-scene swap & 0.0 & 1.38 & 4.32 \\
Single-stage & Zero inputs & 2.0 & 2.68 & 4.98 \\
Two-stage & Normal & 0.0 & 0.00 & 4.38 \\
Two-stage & Cross-scene swap & 27.0 & 4.95 & 7.06 \\
Two-stage & Zero inputs & 33.0 & 8.24 & 9.76 \\
\bottomrule
\end{tabular}
\end{table}

\paragraph{Protocol.} Observations come from 200 navigation-training trajectories in 155 scenes. Fixed single-stage and two-stage models keep images, history, instructions, and other inference settings unchanged; both use full causal attention. Interventions modify the four latent inputs, including the initial hidden state $h_0$; the action-readout input is unchanged. Cross-scene swaps use corresponding inputs from another scene under the same model. Self-input replay matches normal inference bit-for-bit for all 400 model--observation pairs. An action is affected when the first waypoint changes by at least 2\,cm in XY or $2^\circ$ in yaw; 0\% does not mean unchanged outputs. Shift is displacement from normal outputs; ADE is error against the reference trajectory. Both average over eight waypoints from $+0.2$ to $+1.2$\,s. These are open-loop action responses, not closed-loop success or collision rates under intervention. The training schedules differ in both information pathways and loss weights.

\FloatBarrier

\section{Supplementary Analyses}\label{sec:appendix_analysis}

\subsection{Protocols}\label{sec:appendix_protocols}

\paragraph{Table~\ref{tab:ablation}.} All source encoders are frozen; the language model, \mixres where present, and the action head are trained. No row in (a) dreams; every row that dreams uses the free-running rollout of the full model, and each row differs from the full model only in the factors marked. In (b), the w/o-supervision row keeps the rollout but applies neither the prediction loss nor VISReg to the latents, and the native-pooled row predicts the backbone ViT's pooled features without regularization. SPL is on a 0--100 scale.

\paragraph{Readout decoders.} Readouts use lightweight decoders, $\widehat y_r=D_r(h_1,\ldots,h_4)$, trained after the fact on frozen latents to recover geometry, semantics, target states, and motion. They serve only to analyze the representation, and each comparison trains the same decoder on every representation. Figures label the backbone ViT baseline ``Qwen native'' and the w/o-supervision model ``Free run''.

\subsection{\texorpdfstring{The Self-Constructed Space (Section~\ref{sec:analysis_see})}{The Self-Constructed Space (Section 5.1)}}\label{sec:appendix_see}

\begin{figure}[htbp]
\centering
\includegraphics[width=\linewidth]{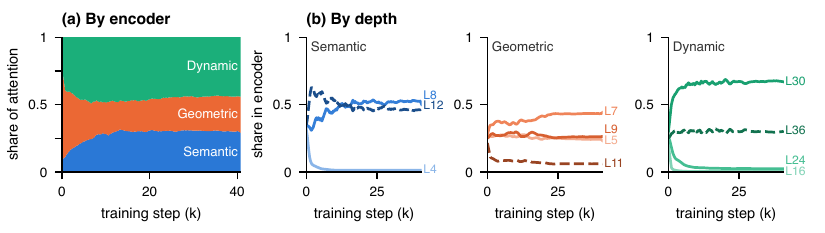}
\caption{\textbf{Attention shares by encoder and by depth.} Attention shares of \mixres during training over a subset of layers spaced along each encoder's depth, averaged over its three layers and all positions. (a) By encoder; the running state, initialized from the geometric encoder, counts as geometric. (b) By depth within each encoder; dashed lines mark final layers.}
\label{fig:alpha_full}
\end{figure}

\paragraph{Attention shares.} Figure~\ref{fig:alpha_full} plots the weights $\alpha$ of Eq.~\ref{eq:fusion_weights} logged during training, averaged over the three layers of \mixres and over positions. The candidates are features from a subset of layers spaced along each encoder's depth, together with the running hidden state, which is initialized from the geometric encoder and therefore counted as geometric. Curves are smoothed with a moving average over 121 logged points, about 1,200 training steps. Table~\ref{tab:layers} varies the number of layers of \mixres.

\paragraph{Preferences by region.} Figure~\ref{fig:alpha}(b) uses the attention weights of \mixres in the final model. Each encoder's weights are summed over its layers, divided by that encoder's spatial mean over the four current views, raised to the fourth power, and normalized across encoders; the running state is excluded. Colors therefore show contrast-enhanced relative preferences, not absolute weights or causal contributions.

\begin{table}[htbp]
\centering\small
\caption{\textbf{Capacity of \mixres is not the source of the gain.} Point-goal navigation without the dream, all three encoders, varying the number of \mixres layers. The default uses three layers (SR 93.45\%, SPL 84.2\%, CR 18.05\%); all variants lie within 2.2 points of one another in SR and SPL, so the gain in Table~\ref{tab:see_ladder} is not a matter of capacity.}
\label{tab:layers}
\begin{tabular}{lccccc}
\toprule
\mixres layers & 1 & 2 & 4 & 8 & 16 \\
\midrule
SR (\%)\,$\uparrow$ & 92.75 & 91.15 & 92.25 & 91.70 & 93.30 \\
SPL (\%)\,$\uparrow$ & 84.1 & 82.6 & 83.0 & 82.6 & 84.1 \\
CR (\%)\,$\downarrow$ & 18.40 & 18.25 & 17.25 & 16.10 & 16.40 \\
\bottomrule
\end{tabular}
\end{table}

\FloatBarrier
\subsection{\texorpdfstring{The Target of the Dream (Section~\ref{sec:analysis_space})}{The Target of the Dream (Section 5.2)}}\label{sec:appendix_space}

\textbf{The regularizer keeps the space from collapsing.} Trained on prediction alone, a joint-embedding predictive architecture collapses, mapping every input to nearly the same representation, which is why it is paired with an anti-collapse regularizer~\citep{maes2026leworldmodel,balestriero2025lejepa}. Our target is no exception: the same \mixres that navigation trains produce it, so it moves as they learn. VISReg constrains only the scale and shape of its distribution~\citep{wu2026visreg}. Without it, success falls from 95.00\% to 92.90\% and collisions rise from 13.45\% to 16.40\%, worse than rolling out the latents with no supervision at all (93.35\% and 15.90\%, Table~\ref{tab:ablation}(b)); SIGReg or VICReg in its place recovers less (Table~\ref{tab:regularizer}).

\begin{table}[htbp]
\centering\small
\caption{\textbf{Choice of the target regularizer.} Same model and data; only the regularizer on the dream targets changes; None keeps the prediction loss and drops only the regularizer.}
\label{tab:regularizer}
\begin{tabular}{l rrr}
\toprule
Regularizer & SR (\%)\,$\uparrow$ & SPL\,$\uparrow$ & CR (\%)\,$\downarrow$ \\
\midrule
None & 92.90 & 84.09 & 16.40 \\
SIGReg & 93.85 & 84.39 & 17.15 \\
VICReg & 93.25 & 83.98 & 16.40 \\
VISReg (ours) & \textbf{95.00} & \textbf{86.70} & \textbf{13.45} \\
\bottomrule
\end{tabular}
\end{table}

\FloatBarrier
\subsection{\texorpdfstring{What the Dream Holds (Section~\ref{sec:analysis_dream})}{What the Dream Holds (Section 5.3)}}\label{sec:appendix_dream}

\begin{figure}[p]
\centering
\includegraphics[width=\linewidth]{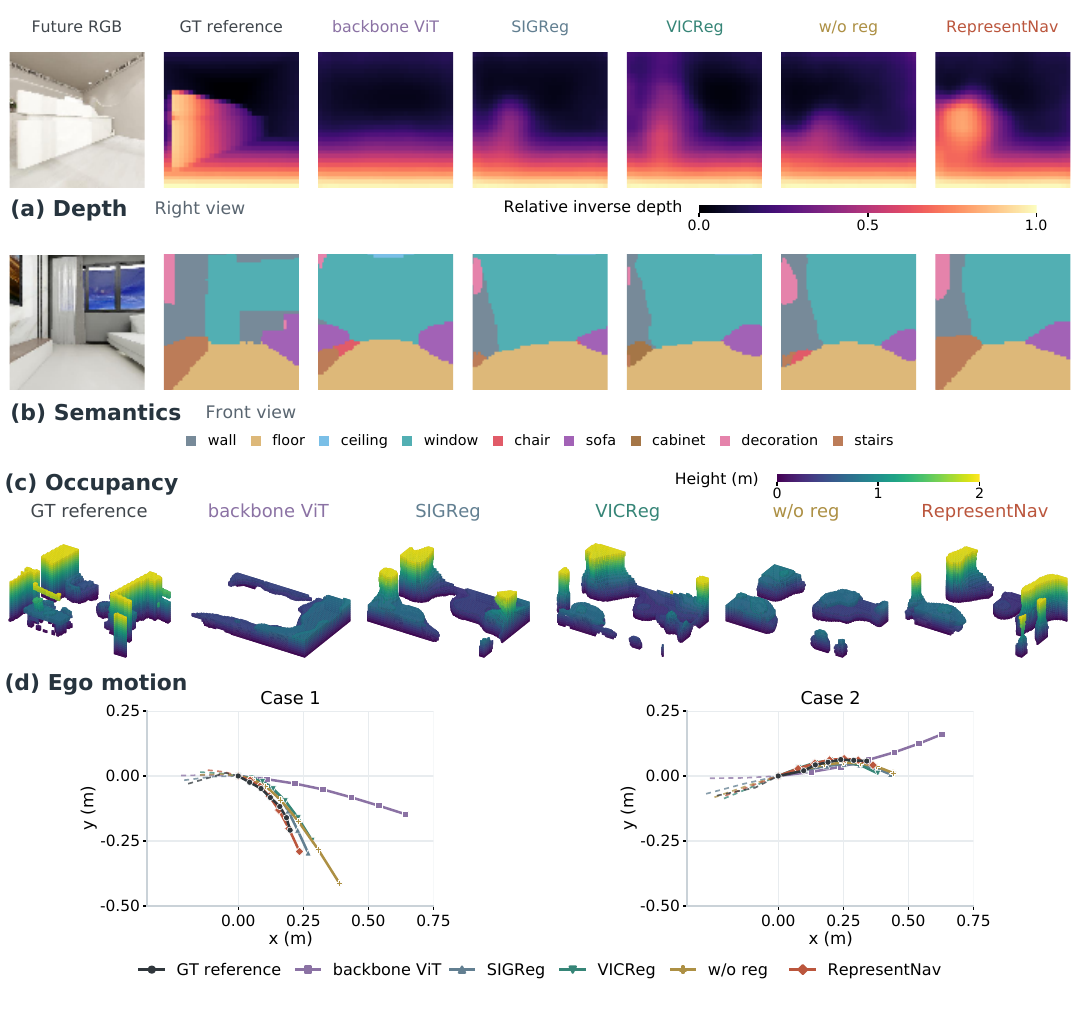}
\caption{\textbf{Future scenes and ego motion can be read from the dream.} Expanded comparison for Figure~\ref{fig:representnav_readouts}, using the same observations and selected camera views. (a) The camera depth uses original $32\times32$ relative inverse-depth maps and GT references on a common color scale. (b) The camera semantics uses original $64\times64$ labels and GT references with a common class palette. Both target $+0.4$\,s; future RGB provides visual context. (c) Occupancy uses native $120\times80\times20$ voxels with identical viewpoints, spatial bounds, and height colors. (d) Ego motion shows six future positions with shared ranges and equal axis scales; dashed and solid curves denote past and future motion. GT sources and full-test metrics are specified in Table~\ref{tab:readouts_mini}. Groups use different observations, selected post hoc to illustrate RepresentNav's strengths. }
\label{fig:representnav_readouts_full}
\end{figure}

Table~\ref{tab:readouts_full} gives the full readout comparison behind Figure~\ref{fig:representnav_readouts}, and Figure~\ref{fig:representnav_readouts_full} expands its examples.

\paragraph{Protocol.} Depth targets are ground-truth depth maps, expressed as normalized relative inverse depth rather than metric depth; occupancy uses native voxel labels. Ego-motion ADE measures XY error at six future positions from $+0.2$ to $+1.2$\,s in the current robot frame. The four pretrained-encoder rows and the \mixres{} [CLS] row encode the real future observation, for openD4RT the four-frame clip ending at it; they are references rather than controls, since they see the frame the latent must predict while the latent sees only the policy's multi-view history. The rows that share the latent's inputs are w/o supervision, w/o regularization, SIGReg, VICReg, and single-stage. RepresentNav uses one fixed two-stage model and raw LLM hidden states. Semantic references are OneFormer-20 pseudo-labels. The test observations span 48 scenes; metrics accumulate over pixels in four $64\times64$ views. Each readout uses one vector per observation: globally pooled pretrained-encoder features, the pre-head [CLS] of \mixres, or one raw LLM latent. LLM latents and backbone ViT features have 2,560 dimensions, and every input vector receives 16 additional static calibration values per camera. Configurations share the semantic decoder, training budget, and validation-selection protocol; results use validation-selected three-seed probability ensembles. Paired scene bootstrap does not establish a clear ordering between single-stage and two-stage semantic readouts.

\begin{table}[htbp]
\centering\footnotesize
\caption{\textbf{Full readout comparison}, including the alternative regularizers and the single-stage model. Each metric uses 1,000 test observations.}
\label{tab:readouts_full}\label{tab:readouts_mini}
\setlength{\tabcolsep}{3pt}
\begin{tabular}{lrrrrr}
\toprule
& Depth & Occupancy & Semantic & Pixel & Ego motion \\
Representation & MAE$\downarrow$ & IoU (\%)$\uparrow$ & mIoU (\%)$\uparrow$ & acc. (\%)$\uparrow$ & ADE (m)$\downarrow$ \\
\midrule
Backbone ViT on real future & 0.16257 & 25.47 & 16.05 & 57.32 & 0.17932 \\
DA3 on real future & 0.14472 & 26.14 & 15.25 & 59.99 & 0.14095 \\
DINOv3 on real future & 0.15206 & 25.85 & 16.84 & 58.72 & 0.14209 \\
openD4RT on real future & 0.10253 & 29.79 & 20.34 & 66.67 & 0.04713 \\
\mixres{} [CLS] on real future & \textbf{0.07520} & \textbf{35.95} & \textbf{24.82} & \textbf{73.45} & \textbf{0.02962} \\
\midrule
w/o supervision & 0.10680 & 31.37 & 18.50 & 67.39 & 0.05149 \\
w/o regularization & 0.10958 & 30.99 & 18.83 & 66.36 & 0.04583 \\
SIGReg & 0.10307 & 32.40 & 21.22 & 69.21 & 0.04539 \\
VICReg & 0.10233 & 31.82 & 22.27 & 69.00 & 0.04687 \\
RepNav, single-stage & 0.09449 & 34.57 & 23.64 & 71.53 & 0.04635 \\
RepNav & 0.09131 & 34.81 & 23.38 & 71.35 & 0.04448 \\
\bottomrule
\end{tabular}
\end{table}

\FloatBarrier

\subsubsection{Target Localization and Motion on Track}
\label{sec:track_readouts}

\begin{figure}[!ht]
\centering
\includegraphics[width=.86\linewidth]{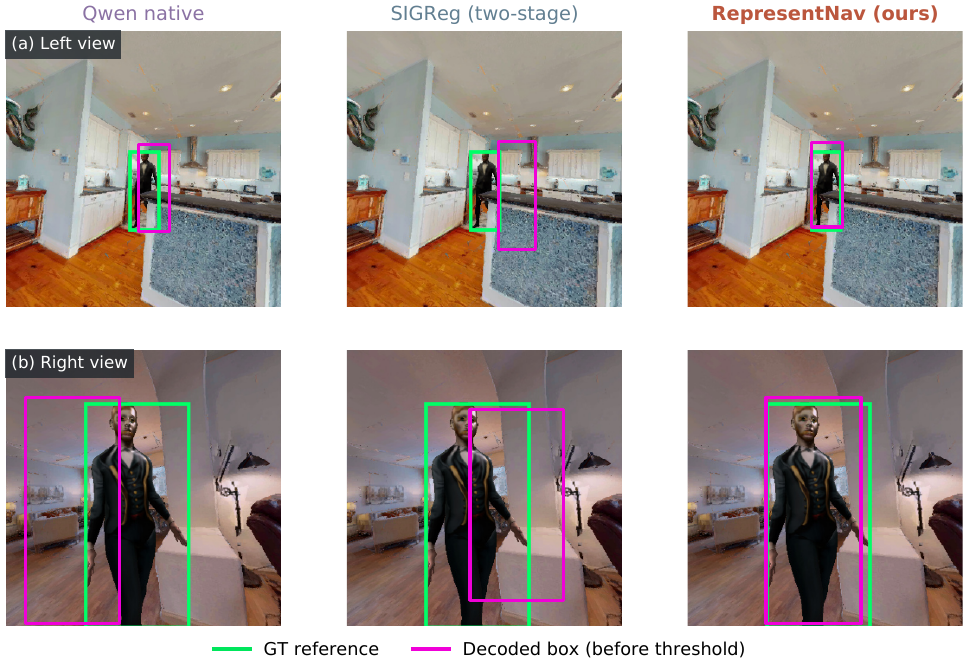}
\caption{\textbf{RepresentNav localizes the target more accurately than the baselines.} Frozen-model bounding-box readouts on two expanded Track test observations. Each row shares the current-view RGB image and GT across models. Green: GT reference. Magenta: original decoded box coordinates before visibility-confidence filtering, shown for all models to compare localization on GT-visible targets, not thresholded detections. RepresentNav's boxes lie closer to GT; the baselines are displaced. Examples are selected post hoc; aggregate results are in Table~\ref{tab:readouts}.}
\label{fig:track_bbox}
\end{figure}

\begin{figure}[!t]
\centering
\includegraphics[width=.90\linewidth]{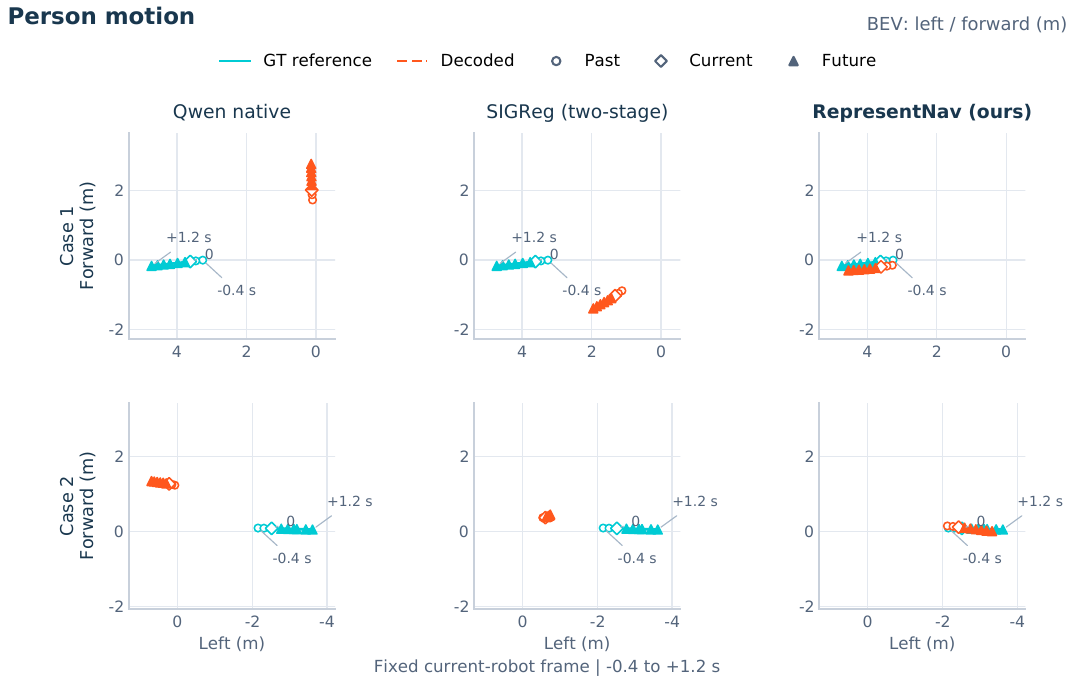}
\caption{\textbf{RepresentNav recovers the target's position and subsequent motion.} Frozen-model readouts on two observations from the expanded Track test set. Each column compares one model with GT: cyan solid curves are GT references; orange dashed curves are readouts. Circles, diamonds, and triangles mark past, current, and future positions. Each row shares metric bounds across models, with left horizontal and forward vertical in the fixed current-robot frame. Examples are selected post hoc; aggregate results and model conventions are in Table~\ref{tab:readouts}.}
\label{fig:track_person}
\end{figure}

\begin{figure}[!t]
\centering
\includegraphics[width=.90\linewidth]{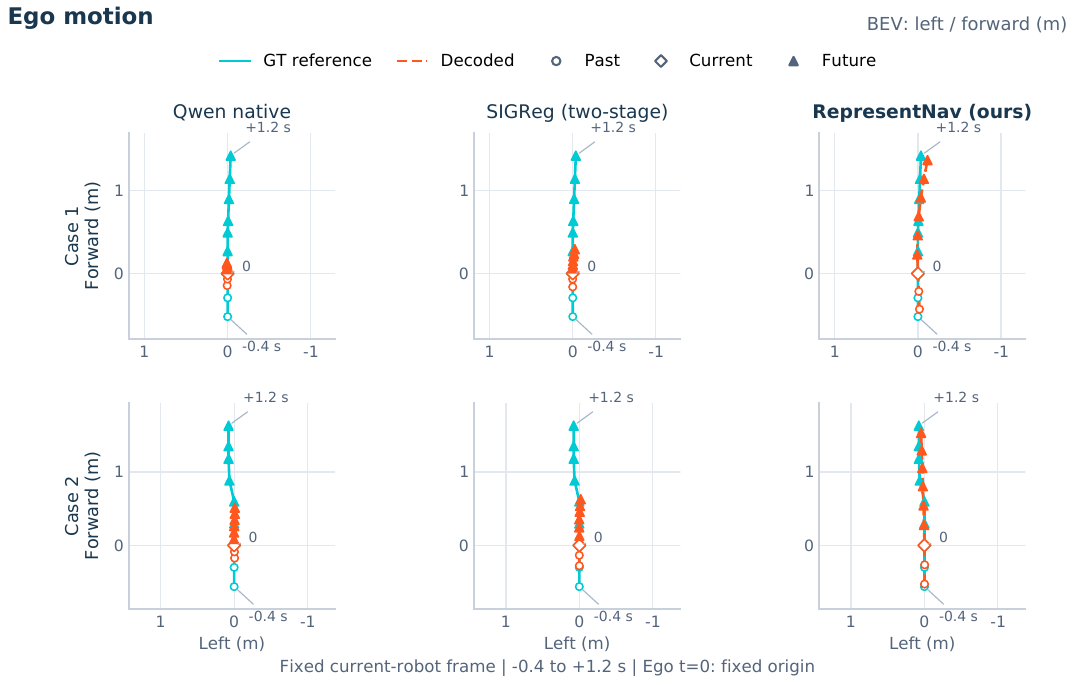}
\caption{\textbf{RepresentNav recovers future ego displacement underestimated by the baselines.} Frozen-model readouts on two observations from the expanded Track test set. Colors, markers, and BEV axes follow Figure~\ref{fig:track_person}; each row shares metric bounds across models. Ego $t=0$ is fixed at the origin. RepresentNav follows the GT displacement, while the backbone ViT and SIGReg predict shorter travel. Examples are selected post hoc. These curves are readouts, not executed trajectories; evaluation conventions are in Table~\ref{tab:readouts}.}
\label{fig:track_ego}
\end{figure}

RepresentNav preserves target location and future motion more accurately on Track than the backbone ViT on the real future images and a two-stage SIGReg model (Table~\ref{tab:readouts}). Current-front box IoU increases from 33.69\% to 59.33\% to 72.81\%, while future person-motion ADE falls from 1.09701\,m to 0.78507\,m to 0.44450\,m. The improvement extends to ego motion, whose future ADE falls from 0.33404\,m to 0.22126\,m to 0.15455\,m. Figures~\ref{fig:track_bbox}--\ref{fig:track_ego} visualize matched Track observations: RepresentNav recovers the target boxes and follows the reference motion more closely in these examples.

\begin{table}[ht]
\centering\small
\caption{\textbf{RepresentNav improves target localization and motion readouts on the expanded Track test set.} All three models are evaluated on the same 5,268 observations.}
\label{tab:readouts}
\label{tab:track_readouts_expanded}
\setlength{\tabcolsep}{4pt}
\resizebox{\linewidth}{!}{%
\begin{tabular}{lccccccc}
\toprule
& \multicolumn{3}{c}{Target localization (\%)} & \multicolumn{2}{c}{Person motion (m)} & \multicolumn{2}{c}{Ego motion (m)} \\
\cmidrule(lr){2-4}\cmidrule(lr){5-6}\cmidrule(lr){7-8}
Model & Front IoU$\uparrow$ & All IoU$\uparrow$ & AP50$\uparrow$ & ADE$\downarrow$ & FDE$\downarrow$ & ADE$\downarrow$ & FDE$\downarrow$ \\
\midrule
Backbone ViT on real future & 33.69 & 32.74 & 8.67 & 1.09701 & 1.27292 & 0.33404 & 0.54533 \\
SIGReg (two-stage) & 59.33 & 50.04 & 36.56 & 0.78507 & 0.90616 & 0.22126 & 0.37249 \\
RepNav (ours) & \textbf{72.81} & \textbf{61.66} & \textbf{69.44} & \textbf{0.44450} & \textbf{0.55498} & \textbf{0.15455} & \textbf{0.25190} \\
\bottomrule
\end{tabular}}
\vspace{3pt}
\begin{minipage}{\linewidth}\footnotesize
\textit{Protocol.} Frozen-model readout evaluation on Track benchmark samples, not closed-loop navigation. The expanded set contains 5,268 observations from eight decoder-test scenes, including additional episode/time anchors and native-camera variants; these are not 5,268 independent trajectories. Front IoU uses GT-visible targets in the current front view. All IoU uses GT-visible targets across four views and nine times from $-0.4$ to $+1.2$\,s; AP50 also accounts for false positives and missed detections. ADE uses future offsets $+0.2$ to $+1.2$\,s; FDE uses $+1.2$\,s. Motion is measured in the fixed current-robot frame. The backbone ViT row encodes the real future images as a reference. RepresentNav uses one fixed two-stage checkpoint. Training configurations differ beyond the regularizer, so this is not an isolated regularizer ablation.
\end{minipage}
\end{table}

\FloatBarrier

\FloatBarrier

\FloatBarrier

\section{Qualitative Results}\label{sec:appendix_qualitative}

\begin{figure}[p]
\centering
\includegraphics[width=\linewidth,height=0.88\textheight,keepaspectratio]{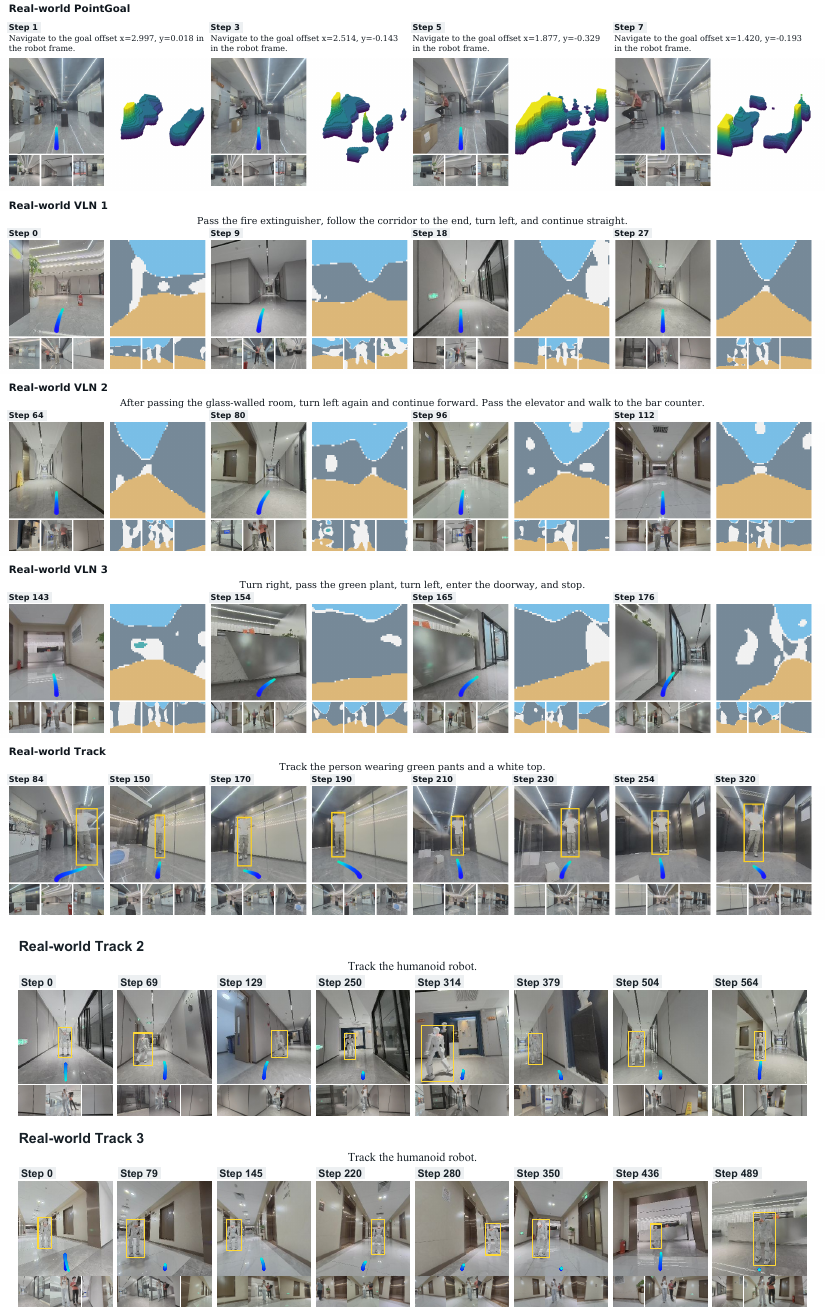}
\caption{\textbf{Visualization of real-world experiments across tasks.} }
\label{fig:realworld_cross_task_visualization}
\end{figure}

Figure~\ref{fig:benchmark_performance_visualization} shows episodes from the benchmarks of Section~\ref{sec:experiments}, and Figure~\ref{fig:realworld_cross_task_visualization} real-world episodes across the three tasks.

\begin{center}
\captionsetup{hypcap=false}
\includegraphics[width=\linewidth,height=0.8\textheight,keepaspectratio]{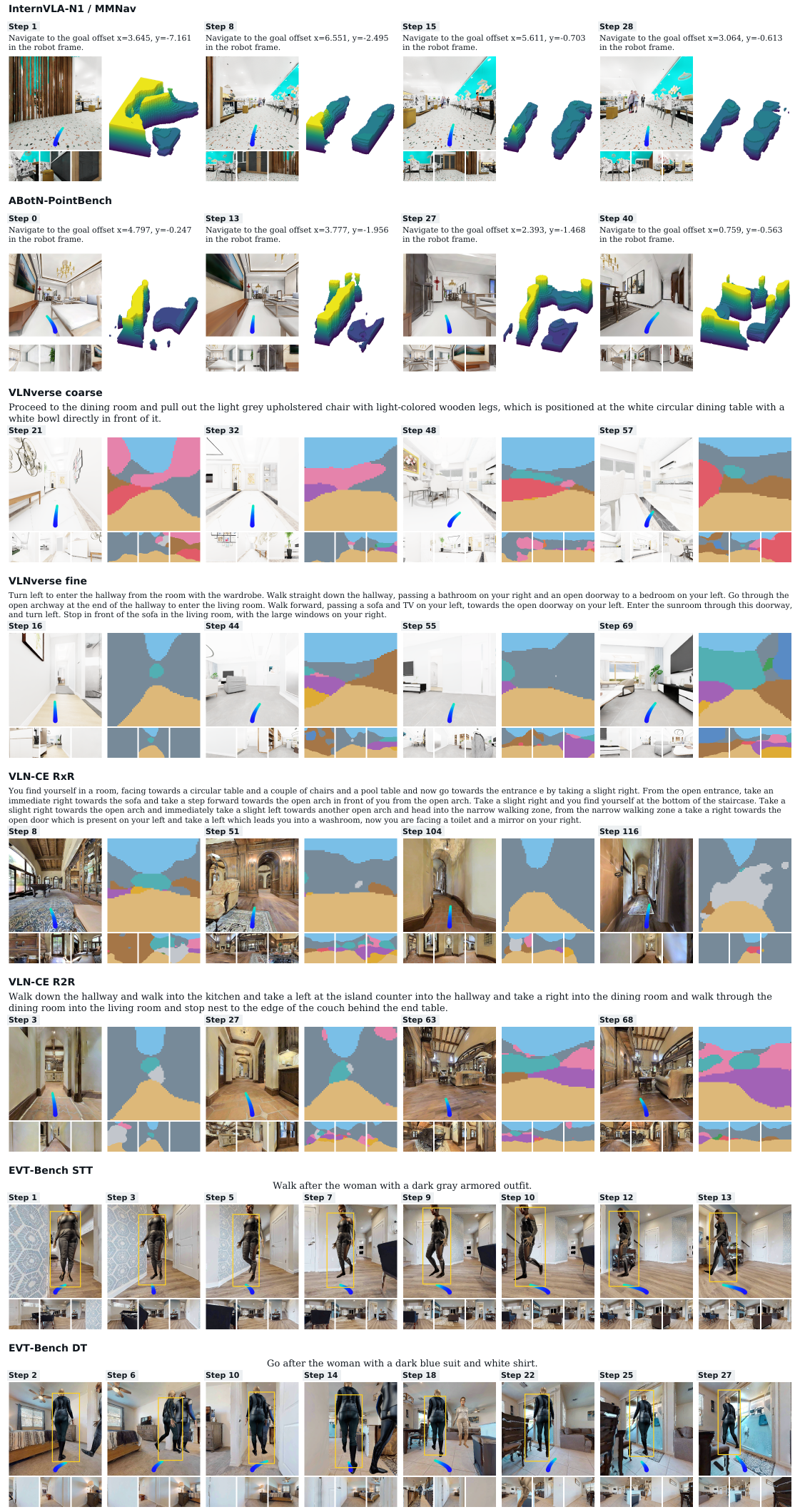}
\captionof{figure}{\textbf{Visualization of performance on benchmarks.}}
\label{fig:benchmark_performance_visualization}\end{center}

\begin{figure}[!htbp]
\centering
\includegraphics[width=0.88\linewidth]{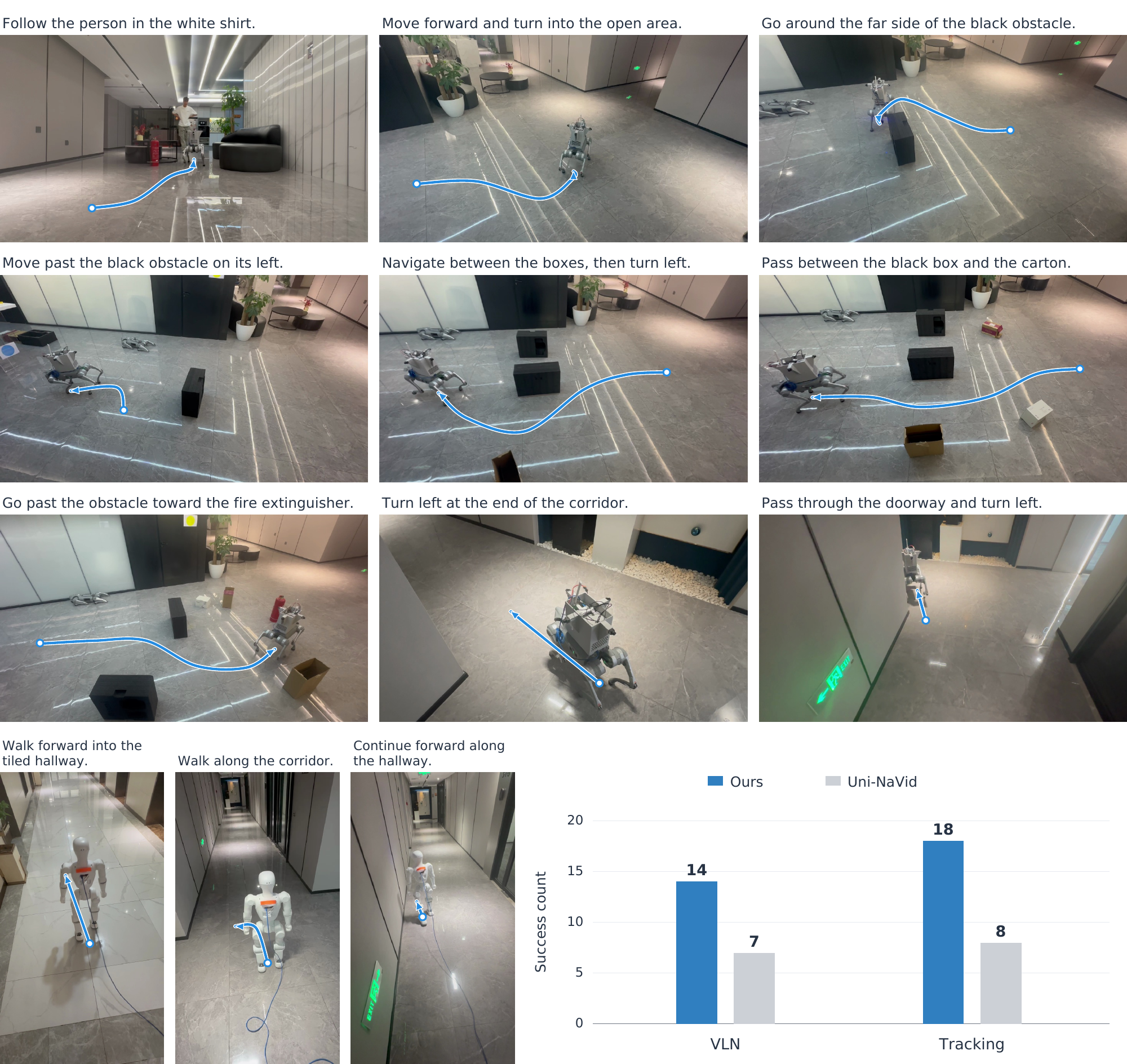}
\caption{\textbf{Third-person real-world demonstrations.} }
\label{fig:realworld_third_person_quantitative}
\end{figure}

\clearpage
\textbf{Real-world performance on 40 test cases.} We evaluate RepresentNav on two robotic platforms, a Go2 quadruped and a Bumi humanoid. Our quantitative evaluation comprises 20 VLN and 20 tracking test cases. VLN requires following natural-language instructions to reach specified destinations, while tracking requires following a designated moving target. Figure~\ref{fig:realworld_third_person_quantitative} presents third-person demonstrations and a quantitative comparison with Uni-NaVid. RepresentNav completes 14 of the 20 VLN cases (70\%) and 18 of the 20 tracking cases (90\%), compared with 7 (35\%) and 8 (40\%) for Uni-NaVid, respectively. These correspond to success-rate improvements of 35 percentage points on VLN and 50 percentage points on tracking.

\textbf{Qualitative real-world demonstrations across tasks.} Figure~\ref{fig:realworld_cross_task_visualization} provides first-person visualizations of point-goal navigation, VLN, and tracking. These examples illustrate how RepresentNav approaches specified goals, follows route instructions through indoor environments, and follows moving targets.

\begin{figure}[!htbp]
\centering
\includegraphics[width=0.82\linewidth,height=0.52\textheight,keepaspectratio]{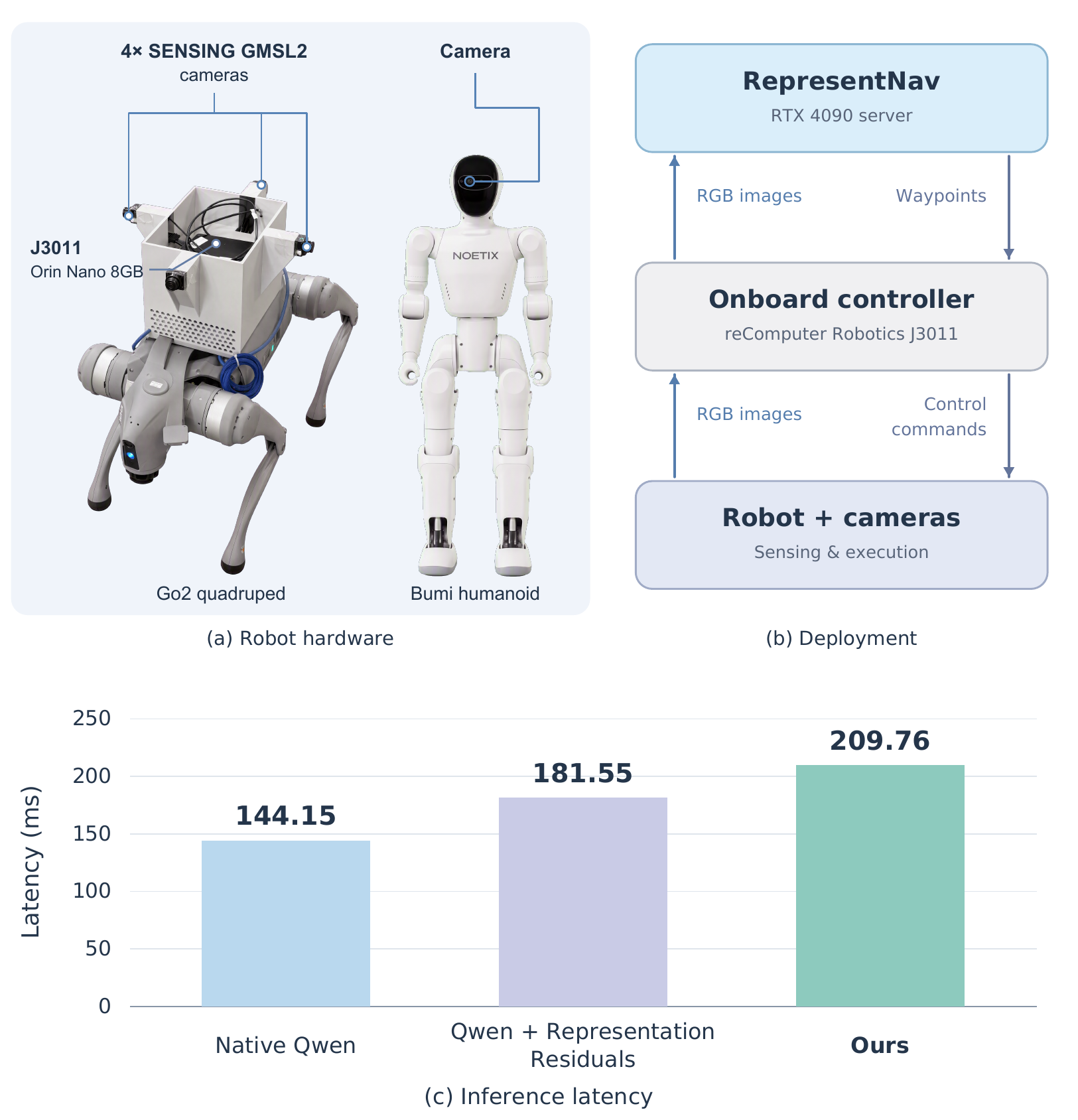}
\caption{\textbf{Real-world robot platform and inference latency.} (a) The Go2 quadruped and Bumi humanoid, with camera locations marked. The Go2 carries four SENSING GMSL2 cameras and a reComputer Robotics J3011 with an NVIDIA Jetson Orin Nano 8GB module. (b) RGB observations are sent to an RTX 4090 server running RepresentNav; predicted waypoints are returned to the onboard controller for robot execution. (c) Inference latency of Native Qwen, Qwen with Representation Residuals, and our full model.}
\label{fig:realworld_robot_platform}
\end{figure}
\FloatBarrier

\FloatBarrier

\FloatBarrier

\section{AI Use Statement}
\label{app:ai_use}
We use Claude to improve the writing of the manuscript. The authors take full responsibility for the final content, analyses, and conclusions.

\section{Ethics Statement}
\label{app:ethics}
The public datasets and simulation assets used in this work are available under licenses permitting research use. Our real-world experiments do not disclose personal information: the faces of people who appear in our recordings are blurred.

\section{Reproducibility Statement}
\label{app:reproducibility}
We describe \mixres in Section~\ref{sec:mixres}, the latent chain-of-thought and its targets in Section~\ref{sec:representation_learning}, the two-stage curriculum in Section~\ref{sec:training}, and the training data in Section~\ref{sec:data}; Section~\ref{sec:experiments} gives the benchmarks, metrics, and implementation details. Appendix~\ref{sec:appendix_training} provides further architecture and training details, including the loss weights and data split of each stage, and Appendix~\ref{sec:appendix_protocols} gives the protocols of the ablations and representation analyses, to support reproducibility.